\documentclass[twocolumns,10pt]{IEEEtran}
\usepackage[T1]{fontenc}

\usepackage{amssymb,amsmath,amsfonts,amsthm}
\usepackage{mathrsfs}
\usepackage{cite}
\usepackage{array}
\usepackage{tabularx}
\usepackage[table]{xcolor}
\usepackage{multicol, multirow}
\usepackage{color}
\usepackage{mathtools}
\usepackage{subcaption}
\usepackage{mathtools}
\usepackage{tikz,pgf}
\usetikzlibrary{calc,positioning,mindmap,trees,decorations.pathreplacing}
\usepackage{graphicx}
\usepackage{bbm}
\usepackage[utf8]{inputenc}
\usepackage{algorithm}

\newtheorem{definition}{Definition}
\newtheorem{theorem}{Theorem}
\newtheorem{lemma}{Lemma}

\newtheorem{remark}{Remark}
\newtheorem{assumption}{Assumption}

\newcommand{\argmin}{\mathop{\rm arg~min}\limits}

\providecommand{\keywords}[1]{\textbf{\textit{Index terms---}} #1}

\IEEEoverridecommandlockouts
\ifCLASSINFOpdf
\else
\fi
\usepackage[cmintegrals]{newtxmath}
\begin{document}
	\title{Scheduling and Aggregation Design for Asynchronous Federated Learning over Wireless Networks}
	
	\author{Chung-Hsuan Hu, Zheng Chen, and Erik G. Larsson
		\thanks{The authors are with the Department of Electrical Engineering (ISY), Link\"{o}ping University, 58183 Link\"{o}ping, Sweden (email: chung-hsuan.hu@liu.se, zheng.chen@liu.se, erik.g.larsson@liu.se).}
	\thanks{This work was supported in part by Zenith, Excellence Center at Link\"{o}ping - Lund in Information Technology (ELLIIT), and the Knut and Alice Wallenberg Foundation.}}
	
	\maketitle
	
\begin{abstract}	
Federated Learning (FL) is a collaborative machine learning (ML) framework that combines on-device training and server-based aggregation to train a common ML model among distributed agents. In this work, we propose an asynchronous FL design with periodic aggregation to tackle the straggler issue in FL systems. Considering limited wireless communication resources, we investigate the effect of different scheduling policies and aggregation designs on the convergence performance. Driven by the importance of reducing the bias and variance of the aggregated model updates, we propose a scheduling policy that jointly considers the channel quality and training data representation of user devices.
The effectiveness of our channel-aware data-importance-based scheduling policy, compared with state-of-the-art methods proposed for synchronous FL, is validated through simulations.
Moreover, we show that an ``age-aware'' aggregation weighting design can significantly improve the learning performance in an asynchronous FL setting.
\end{abstract} 

\keywords{Federated Learning, asynchronous training, wireless networks, scheduling, aggregation}

\section{Introduction}
The training of machine learning (ML) models usually requires a massive amount of data. Nowadays, the ever-increasing number of connected user devices has benefited the development of ML algorithms by providing large sets of data that can be utilized for model training. As privacy concerns become vital in our society, using private data from user devices for training ML models becomes tricky. Therefore, federated learning (FL) with on-device information processing has been proposed for its advantages in preserving data privacy. FL is a collaborative ML framework where multiple devices participate in training a common global model based on locally available data \cite{konevcny2016federated}. Unlike centralized ML architecture wherein the entire set of training data need to be centrally stored, in an FL system, only model parameters are shared between user devices and a parameter server. Due to the heterogeneity of participating devices, local training data might be unbalanced and not independent and identically distributed (i.i.d.), which deviates FL from conventional distributed optimization frameworks where homogeneous and evenly distributed data are assumed.

Federated Averaging (FedAvg) is one of the most representative and baseline FL algorithms \cite{mcmahan2017communication}, with an iterative process of model broadcasting, local training, and model aggregation. In every iteration, the model aggregation process can start only when all the devices have finished local training. Thus, the duration of one iteration is seriously limited by the slowest device \cite{chen2016revisiting}. This phenomenon, commonly observed in synchronous FL methods, is known as the straggler issue.
One approach to this issue is altering the synchronous procedure to an asynchronous one, i.e., the server does not need to wait for all the devices to finish local training before conducting updates aggregation. In the literature, such an asynchronous FL framework has been adopted in many deep-learning settings \cite{xie2019asynchronous,chen2020asynchronous}. However, fully asynchronous FL with sequential updating \cite{xie2019asynchronous,chen2020asynchronous} can lead to high communication costs due to frequent model exchanges.
Hence, we propose an asynchronous FL framework with periodic aggregation, which eliminates the straggler effect without excessive model updating and information exchange between the server and participating devices. As compared to other existing works on FL with asynchronous updates \cite{wang2021resource,Nguyen2022Fed,zhou2021TeaFed}, our proposed design is easy to implement and requires a small amount of side information.

Communication resource limitation is another critical issue in wireless FL systems. As model exchanges take place over wireless channels, the system performance (communication costs and latency) naturally suffers from the limitation of frequency/time resources, especially when the number of participating devices is large. One possible solution to reduce the communication load is to allow a fraction of participating devices to upload their local updates for model aggregation. Then, depending on the allocated communication resources and wireless link quality, each device compresses its model updates accordingly such that the compressed updates can be transmitted reliably given the allocated resources. 
Device scheduling and communication resource allocation are critical for achieving communication-efficient FL over wireless networks \cite{gafni2021federated, Yang2020scheduling}. Intuitively, 
devices with a higher impact on the learning performance should be prioritized in scheduling. This learning-oriented communication objective stands in contrast to the conventional rate-oriented design adopted in cellular networks, where the aim is to achieve higher spectral efficiency or network throughput. 
Several existing works consider different metrics to indicate the significance of local updates, such as norm of model updates \cite{amiriconvergence, luo2022tackling} and Age-of-Update (AoU) \cite{yang2019agebased}. Under the motivation of receiving model updates with less compression loss, some works consider wireless link quality in scheduling design \cite{salehi2021federated, importance-aware, Ozfatura2021fastF,chen2022federated, luo2022tackling}. On the other hand, to take into account the non-i.i.d. data distribution, uncertainty of data distribution is considered in \cite{importance-aware}, and in \cite{taik2021data, shen2022variance} the scheduling design follows the principle of giving higher priority to devices with larger diversity in their local data. Some works consider joint optimization of device scheduling and resource allocation towards minimal latency \cite{shi2021jointD} or empirical loss \cite{wadu2021jointC,chen2020aJoint}.  
Nevertheless, all of them consider synchronous FL systems. Few existing works have considered the design in an asynchronous setting. In \cite{Lee2021adaptive}, scheduling in asynchronous FL is considered based on maximizing the expected sum of training data subject to the uncertainty of channel conditions, data arrivals, and limited communication resources. However, the effect of non-i.i.d. data distribution is not considered in the scheduling design. 

Compared to the synchronous setting, asynchronous FL needs to deal with the asynchrony of local model updates since different devices may perform local training based on different versions of the global model.  
Some heuristic aggregation designs are explored in the literature. 
In \cite{Lee2021adaptive}, devices with more frequent transmission failures from past iterations will transmit enlarged gradient updates. In \cite{chai2020fedat}, larger weights are given to slower tiers in the aggregation process because slower tiers contribute less frequently to the global model.
Both approaches aim at equalizing the contributions from different devices, though with i.i.d. data, it might slow down the convergence since model updates obtained from older global models might contain little useful information to the current version.

We summarize the main contributions of this work:
\begin{itemize}
	\item We propose an asynchronous FL framework with periodic aggregation that achieves fast convergence in the presence of stragglers, and avoids excessive model updating as in fully asynchronous FL settings.
	\item We propose a scheduling policy that jointly considers the channel quality and the training data distribution, aiming at reducing the variance and bias of the aggregated model updates. The effectiveness of the proposed method is supported both by theoretical convergence analysis and simulation results.
	\item We propose an age-aware weighting design for model aggregation to mitigate the effects of update asynchrony.
	\item We highlight the impact of update compression, data heterogeneity, and intra-iteration asynchrony in the convergence analysis, which also provides a theoretical motivation for our scheduling design.  
\end{itemize}

\section{System Model}
\label{sec:system_model}
We consider an FL system with $N$ devices participating in training a shared global learning model, parameterized by a $d$-dimensional parameter vector $\boldsymbol{\theta}\in\mathbb{R}^d$. Denote $\mathcal{N}=\{1,...,N\}$ as the device set in the system. Each device $k\in\mathcal{N}$ holds a set of local training data $\mathcal{S}_k$ with size $|\mathcal{S}_k|$. Let $\mathcal{S}=\cup_{k\in\mathcal{N}}\mathcal{S}_k$ represent the entire data set in the system with size $|\mathcal{S}|$, where $\mathcal{S}_i\cap \mathcal{S}_j=\emptyset$, $\forall i\neq j$.  The objective of the system is to find the optimal parameter vector $\boldsymbol{\theta}^*$ that minimizes an empirical loss function defined by
\begin{equation}
	F(\boldsymbol{\theta})=\frac{1}{|\mathcal{S}|}\sum_{x\in\mathcal{S}}l(\boldsymbol{\theta},x),
	\label{eq:globalLoss}
\end{equation}
where $l(\boldsymbol{\theta},x)$ is a sample-wise loss function computed based on the data sample $x$. Similarly, a local loss function of device $k$ is defined as 
\begin{equation*}
	F_k(\boldsymbol{\theta})=\frac{1}{|\mathcal{S}_k|}\sum_{x\in\mathcal{S}_k}l(\boldsymbol{\theta},x).
\end{equation*}
Then, we can rewrite $\eqref{eq:globalLoss}$ as
\begin{equation}
	F(\boldsymbol{\theta})=\sum_{k\in\mathcal{N}}\frac{|\mathcal{S}_k|}{|\mathcal{S}|}F_k(\boldsymbol{\theta}).
	\label{eq:globalLoss_f}
\end{equation}     
To characterize the heterogeneity of data distribution in $\mathcal{N}$, we define a metric \cite{li2019convergence}
\begin{equation}
	\Gamma=F^*-\sum_{k\in\mathcal{N}}\frac{|\mathcal{S}_k|}{|\mathcal{S}|}F_k^*,
	\label{def:Gamma}
\end{equation}
where $F^*=\underset{\boldsymbol{\theta}}{\mbox{min}}\,F(\boldsymbol{\theta})$ and $F_k^*=\underset{\boldsymbol{\theta}}{\mbox{min}}\,F_k(\boldsymbol{\theta})$ are the minimal global and local loss, respectively. With i.i.d. data distribution over devices, $\Gamma$ asymptotically approaches $0$ when $|\mathcal{S}|$ increases, while in non-i.i.d. scenario we expect $\Gamma\neq0$, which reflects the data heterogeneity level among devices.

\subsection{FedAvg with Synchronous Training and Aggregation}

\begin{figure}[t!]
	\centering
	\includegraphics[width=\columnwidth]{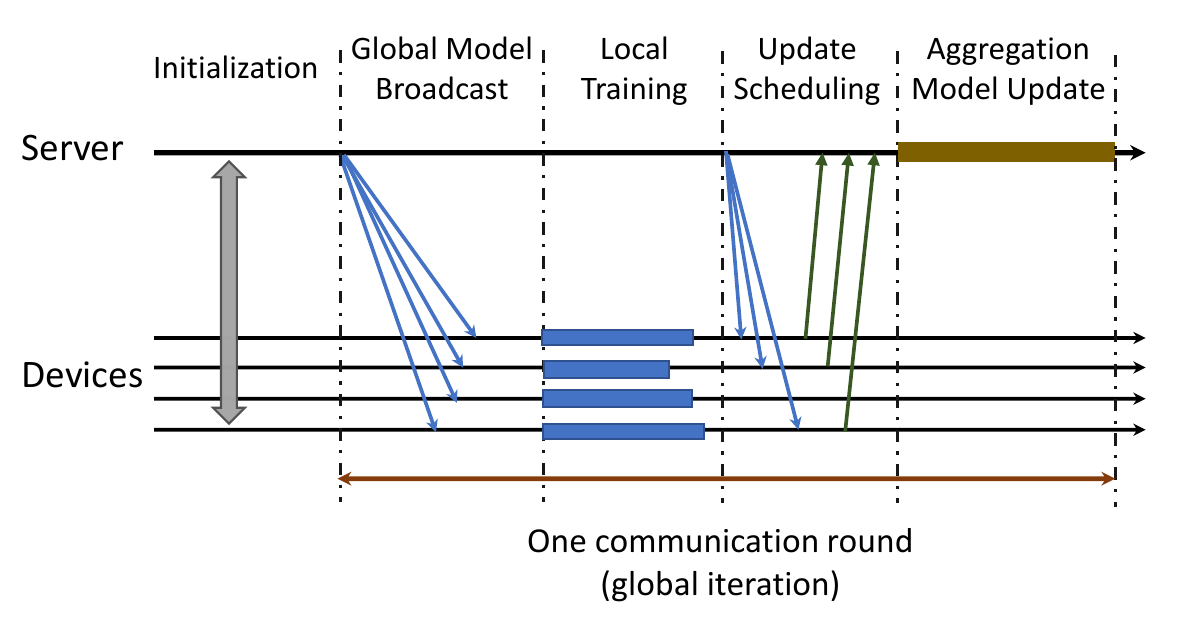}
	\caption{The FL process and information exchange between the server and the participating devices.}
	\label{fig:FL}
\end{figure}
FedAvg is one representative FL algorithm with synchronous procedure of local training and global aggregation. The entire training process is divided into many global iterations (communication rounds), where during each iteration, the server aggregates the model updates from the participating devices computed over their locally available data. 
We consider a modified version of FedAvg, where an extra step of device scheduling is added after local training, as illustrated in Fig.~\ref{fig:FL}. The main motivation behind this is to reduce the communication costs and delay, especially in a wireless network with limited communication resources. In the $t$-th global iteration, $t=1,2,\ldots$, the following steps are executed:
\begin{enumerate}
	\item The server broadcasts the current global model $\boldsymbol{\theta}(t)$ to the device set $\mathcal{N}$.
	\item Each device $k\in \mathcal{N}$ runs $E$ steps of stochastic gradient descent (SGD). The corresponding update rule follows
	\begin{equation*}
		\boldsymbol{\theta}_{k}(t,\tau+1)=\boldsymbol{\theta}_{k}(t,\tau)-\alpha(t,\tau)\nabla F_k(\boldsymbol{\theta}_{k}(t,\tau);\mathcal{B}_k(t,\tau)),
	\end{equation*}
	where $\tau=0,...,E-1$ indicates the local iteration index, $\boldsymbol{\theta}_{k}(t,0)=\boldsymbol{\theta}(t)$, $\alpha(t,\tau)$ represents the learning rate and $\nabla F_k(\boldsymbol{\theta}_{k}(t,\tau);\mathcal{B}_k(t,\tau))$ denotes the gradient computed based on a randomly selected mini-batch $\mathcal{B}_k(t,\tau)\subseteq\mathcal{S}_k$.
	After completing the local training, each device obtains the model update as the difference between the model parameter vector before and after training, i.e.,
	\begin{equation}
		\boldsymbol{u}_k(t)=\boldsymbol{\theta}_{k}(t,E)-\boldsymbol{\theta}_{k}(t,0)
		\label{eq:local_update}
	\end{equation}
	\item After local training, a subset of devices $\Pi(t)\subseteq\mathcal{N}$ are scheduled for uploading their model updates to the server.\footnote{This uplink scheduling step is particularly important for FL over wireless networks as communication resources need to be shared among devices.}
	\item After receiving the local updates from the scheduled devices, the server aggregates the received information and updates the global model according to
	\begin{equation}
		\boldsymbol{\theta}(t+1)=\boldsymbol{\theta}(t)+\sum_{k\in\Pi(t)}\frac{|\mathcal{S}_k|}{\sum_{j\in\Pi(t)}|\mathcal{S}_j|}\boldsymbol{u}_k(t).
		\label{eq:syncFlAggregation}	
	\end{equation}
\end{enumerate}
This iterative procedure continues until convergence.
\label{sec:fedavg}  

\subsection{Asynchronous FL with Periodic Aggregation}
\begin{figure}
	\centering
	\includegraphics[width=\columnwidth]{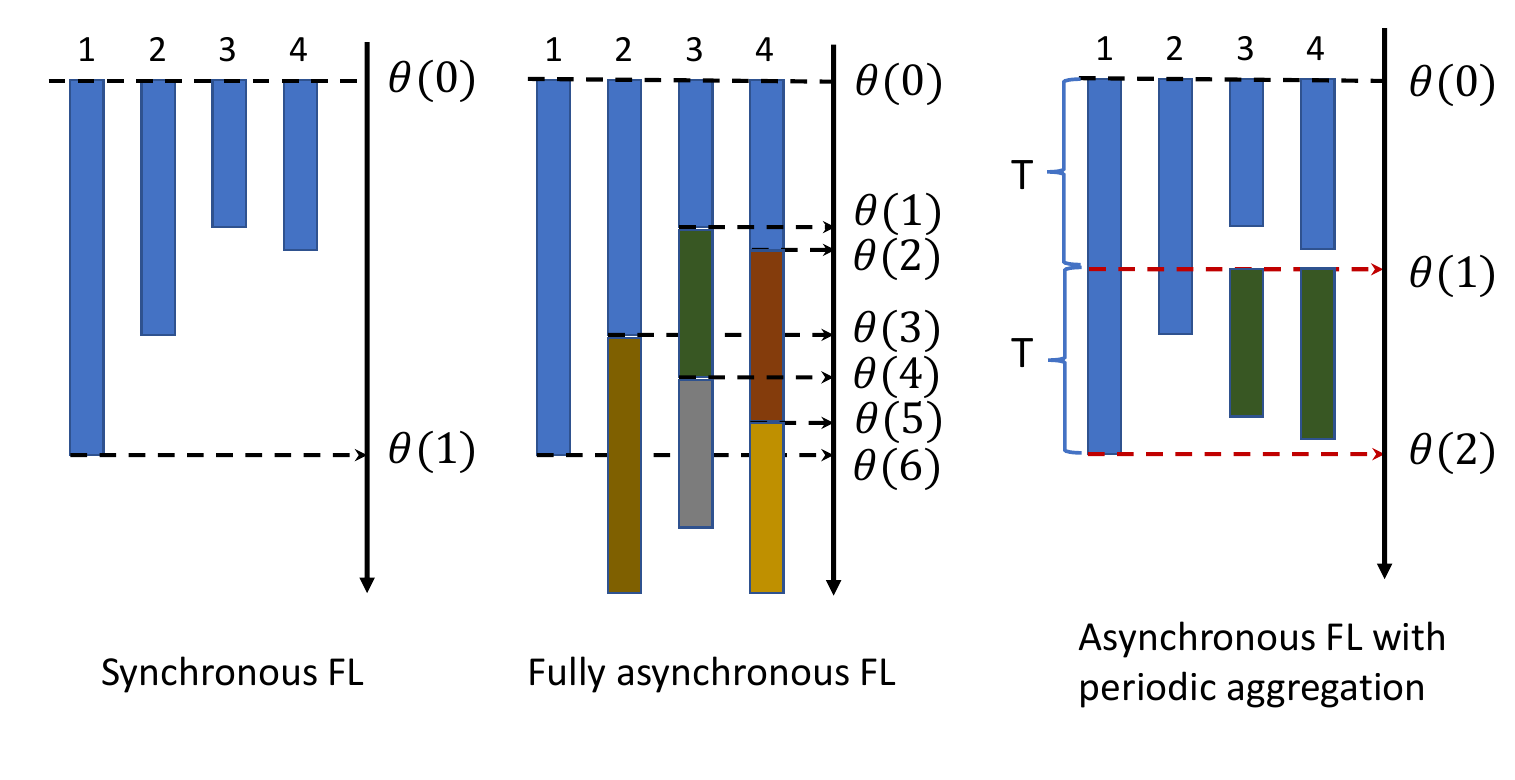}
	\caption{Conceptional difference between synchronous FL, fully asynchronous FL, and our proposed asynchronous FL with periodic aggregation. $\boldsymbol{\theta}(t)$ represents the model parameter vector in the $t$-th global iteration.}
	\label{fig:FL-syn-asyn}
	\vspace{-.2cm}
\end{figure}

To address the straggler issue in synchronous FL without generating excessive communication load, we propose an asynchronous FL framework with periodic aggregation. The general idea is to allow asynchronous training at different devices, with the server periodically collecting updates from those devices that have completed their computation, while the rest continue their local training without being interrupted or dropped. Fig.~$\ref{fig:FL-syn-asyn}$ shows an example of the training and updating timeline of the original synchronous FL, fully asynchronous FL \cite{xie2019asynchronous,chen2020asynchronous}, and our proposed scheme.
Since different devices have different computing capabilities, whenever a device finishes its local training it sends a signal to the server indicating its readiness for update reporting. At every time duration $\tilde{T}$, the server schedules a subset of ready-to-update devices.
We define $\mathcal{K}(t)$ as the set of ready-to-update devices in the $t$-th global iteration. Let $\Pi(t)\subseteq \mathcal{K}(t)$ be the set of scheduled devices, with $|\Pi(t)|=\min\{R,|\mathcal{K}(t)|\}$, where $R$ is the maximal number of devices that can be scheduled. We adopt a regularization term in the loss function to alleviate potential model imbalance caused by asynchronous training. The regularized local loss at device $k$ in the $t$-th global iteration and $\tau$-th local iteration is defined by
\begin{equation*}
	H_k(t,\tau) = F_k\left(\boldsymbol{\theta}_{k}(t,\tau)\right)+\frac{\lambda}{2}\|\boldsymbol{\theta}_{k}(t,\tau)-\boldsymbol{\theta}_{k}(t,0)\|_2^2,
\end{equation*}
where $\lambda>0$ is the regularization coefficient. For any device $k\in \Pi(t)$, its local update is thus computed as
\begin{equation}
	\boldsymbol{\theta}_{k}(t,\tau+1)=\boldsymbol{\theta}_{k}(t,\tau)-\alpha(t,\tau)\nabla H_k(t,\tau),
	\label{eq:localItUpdate_async}
\end{equation}
with $\tau=0,...,E-1$ and $\nabla H_k(t,\tau)$ evaluated on a randomly selected mini-batch $\mathcal{B}_k(t,\tau)\subseteq\mathcal{S}_k$. 
Due to the asynchronous setup, the initial model in the first local iteration is
\begin{equation*}
	\boldsymbol{\theta}_{k}(t,0)=\boldsymbol{\theta}(s_k(t)),
\end{equation*}
where
\begin{equation*}
	s_k(t)=\underset{t'<t}{\mbox{max}}\{t'|k\in \mathcal{K}(t')\}+1
\end{equation*} 
indicates the latest global iteration in which device $k$ received an updated global model.
After receiving the updates from all devices in $\Pi(t)$, the server conducts model aggregation as
\begin{equation*}
	\boldsymbol{\theta}(t+1)=\sum_{k\in\Pi(t)}w_k(t)\left[\boldsymbol{\theta}_k(t,0)+\boldsymbol{u}_{k}(t)\right],	
\end{equation*}
where $\boldsymbol{u}_{k}(t)$ is defined in \eqref{eq:local_update}, and $w_k(t)$ denotes the weight coefficient, with $\sum_{k\in\Pi(t)}w_k(t)=1$. The updated model is then broadcast to all the devices in $\mathcal{K}(t)$ as the new global model to continue local training.
Inspired by the concept of Age of Information (AoI) \cite{kosta2017age}, we define the Age of Local Update (ALU) as
\begin{equation}
	a_k(t)=t-s_k(t),
	\label{eq:definition-alu}
\end{equation}
which represents the elapsed time since the last reception of an updated global model.\footnote{Note that this age-based definition is different from the Age of Update (AoU) proposed in \cite{yang2019agebased}, which measures the elapsed time at each device since its last participation in model aggregation.} The weight $w_k(t)$ can be related to the training data size $|\mathcal{S}_k|$, or the ALU $a_k(t)$, or combination of the both.
Since the transmission of the model updates $\boldsymbol{u}_{k}(t), \forall k\in\Pi(t)$ is subject to communication resource constraints, a compressed version of $\boldsymbol{u}_{k}(t)$, denoted as $\hat{\boldsymbol{u}}_{k}(t)$, will be transmitted over the wireless channels. The compression scheme will be elaborated in the next sub-section.
\label{sec:ayncFL}

\subsection{Physical Layer (PHY) Model}
In an FL process, the communication part takes place in two phases, the server broadcasting global model to the devices (downlink transmission) and the devices reporting model updates to the server (uplink transmission). In general, the server has much higher transmit power than the devices, and the downlink transmission does not require dividing communication resources due to the broadcast channel. Therefore, we assume no compression errors in the downlink. However, uplink transmission suffers from the communication bottleneck, which makes the scheduling and resource allocation design particularly important.

We consider that at any global iteration $t$ the model updates from all devices are transmitted over a block fading channel with $n$-symbol coherence block. We assume orthogonal resource allocation among devices, i.e., $n_k(t)$ symbols are exclusively allocated to the device $k$ and $\sum_{k=1}^{|\Pi(t)|}n_k(t)=n$. Therefore, the  transmissions from multiple devices are interference-free. 
Let $\beta_k$ and $h_k(t)$ denote the large-scale and small-scale fading of the channel from the $k$-th device to the server, respectively. Assuming that the $k$-th device has transmit power $P_{k}(t)\leq P_{k}^{\max}$ and the additive noise in the channel follows $CN(0,\sigma_w^2)$, the channel capacity of the $k$-th link is
\begin{equation}
	C_k(t)=\log_2\left(1+\frac{P_{k}(t)\beta_k|h_k(t)|^2}{\sigma_w^2}\right).
	\label{eq:cpct}
\end{equation}
We consider that the devices apply appropriate data compression and channel coding scheme according to the channel capacity in \eqref{eq:cpct} such that the transmissions of the model updates are error-free. Consequently, the server can reliably receive up to $n_k(t)C_k(t)$ bits from device $k$. The symbol allocation, $n_k(t),\forall k$, and data compression scheme are specified below.  
\subsubsection{Communication resource allocation}
To achieve the same level of compression loss in model updates from each device, we allocate the symbol resources in a way that 
\begin{equation}
	n_1(t)C_1(t)=...=n_{|\Pi(t)|}(t)C_{|\Pi(t)|}(t),
	\label{eq:nc}
\end{equation}
which implies that the devices with better channels are allocated with less symbols. Similar design is considered in \cite{amiriconvergence}.

\subsubsection{Sparsification and Quantization}
Each device needs to compress $\boldsymbol{u}_k(t)$ according to the budget $n_k(t)C_k(t)$ bits.
We utilize random sparsification followed by low-precision random quantization for model compression. Specifically, we define a sparsified model update $\tilde{\boldsymbol{u}}_k(t)\in\mathbb{R}^d$, where the $i$-th component $\tilde{u}_{k,i}(t), \forall i=1,...,d$ is defined by
\begin{equation*}
	\tilde{u}_{k,i}(t)=\begin{cases}
		u_{k,i}(t), & i\in\mathcal{V}_k(t)\\
		0, & \text{otherwise.}
	\end{cases}
\end{equation*}
Here, $u_{k,i}(t)$ is the $i$-th component of $\boldsymbol{u}_k(t)$, and $\mathcal{V}_k(t)$ denotes the set of reserved elements with $|\mathcal{V}_k(t)|=r_k(t)$. Then, each element  $\tilde{u}_{k,i}(t), \forall i\in\mathcal{V}_k(t)$ is processed with $\nu$-level random quantizer $\mathcal{Q}_\nu:\mathbb{R}\rightarrow\mathbb{R}$ such that \cite{Alistarh2017QSGDCS}
\begin{equation*}
	\mathcal{Q}_\nu(\tilde{u}_{k,i}(t))=||\tilde{\boldsymbol{u}}_k(t)||_2\cdot \text{sign}(\tilde{u}_{k,i}(t))\cdot\mathcal{Z}_{\nu}(\tilde{u}_{k,i}(t),||\tilde{\boldsymbol{u}}_k(t)||_2),
\end{equation*}
where
\begin{equation*}
	\mathcal{Z}_{\nu}(u,||\boldsymbol{u}||_2)=\begin{cases}
		\frac{z+1}{\nu}, & \text{with probability $\frac{\nu|u|}{||\boldsymbol{u}||_2}-z$}\\
		\frac{z}{\nu}, & \text{otherwise}\\
	\end{cases}
\end{equation*}
is a random variable, $z=\lfloor\frac{\nu|u|}{||\boldsymbol{u}||_2}\rfloor$, and $\nu$ is the quantization level.\footnote{If $\boldsymbol{u}$ is a zero vector, all the quantized elements are zero.} The resulted compressed vector $\hat{\boldsymbol{u}}_k(t)$ has the components $\hat{u}_{k,i}(t)=\boldsymbol{1}\{i\in \mathcal{V}_k(t)\}\cdot \mathcal{Q}_\nu(\tilde{u}_{k,i}(t)), \forall i$. This random quantizer is unbiased, i.e., $\mathbb{E}[\hat{\boldsymbol{u}}_k(t)]=\tilde{\boldsymbol{u}}_k(t)$, and has quantitative bounded variance $\mathbb{E}[||\hat{\boldsymbol{u}}_k(t)-\tilde{\boldsymbol{u}}_k(t)||_2^2]\leq\frac{r_k(t)||\tilde{\boldsymbol{u}}_k(t)||_2^2}{4\nu^2}$.
Under a fixed quantization level, we find the maximal $r_k(t)$ that satisfies the transmission constraint
\begin{equation}
	\log_2{d\choose r_k(t)}+32+r_k(t)\left(\left\lceil \log_2(\nu+1)\right\rceil+1\right)\leq n_k(t)C_k(t).
	\label{eq:bitQuota}
\end{equation}
Here, the first term $\log_2{d\choose r_k(t)}$ bits are used for sending the indexes $\mathcal{V}_k(t)$, the second term $32$ bits are for the vector norm value, and the third term means that $\left\lceil \log_2(\nu+1)\right\rceil+1$ bits are needed for transmitting each non-zero element of $\hat{\boldsymbol{u}}_k(t)$.

\subsection{Motivation of Scheduling and Aggregation Design}
\label{subsec:prob_form}
Under the proposed asynchronous FL setup, the key design questions are:
\begin{enumerate}
	\item Given the communication resource constraints, how should we schedule a subset of $\mathcal{K}(t)$ for model aggregation under the scenario of heterogeneous training data distribution and wireless link quality?  
	\item Different devices might have different ALUs, either more recent or more outdated. How to design an appropriate weighting policy taking into account the freshness of model updates?   
\end{enumerate}
Recall that the goal is to minimize the global empirical loss $F(\boldsymbol{\theta})$ given in \eqref{eq:globalLoss_f}. We define the optimal model as $\boldsymbol{\theta}^*=\mbox{argmin}_{\boldsymbol{\theta}}\,F(\boldsymbol{\theta})$.
Assuming $F(\boldsymbol{\theta})$ is strongly convex, implying $\nabla F\left(\boldsymbol{\theta}^*\right)=\boldsymbol{0}$, and $L$-smooth, i.e., for any $\boldsymbol{\theta}_1,\boldsymbol{\theta}_2\in\mathbb{R}^d$,
\begin{equation}
	F(\boldsymbol{\theta}_1)-F(\boldsymbol{\theta}_2)\leq\nabla F(\boldsymbol{\theta}_2)^T(\boldsymbol{\theta}_1-\boldsymbol{\theta}_2)+\frac{L}{2}||\boldsymbol{\theta}_1-\boldsymbol{\theta}_2||_2^2,
	\label{asumpt:L-smooth}
\end{equation}
at the $(t+1)$-th iteration we have
\begin{equation}
	\mathbb{E}\left[F\left(\boldsymbol{\theta}\left(t+1\right)\right)\right]-F^*\leq \frac{L}{2}\mathbb{E}\left[||\boldsymbol{\theta}\left(t+1\right)-\boldsymbol{\theta}^*||_2^2\right],
	\label{eq:opt_localSGD}
\end{equation}
where $\mathbb{E}\left[\cdot\right]$ denotes the expectation over all the randomness in the past iterations $1,...,t$. Note that \eqref{eq:opt_localSGD} quantifies the effectiveness of model training by examining the gap between $\mathbb{E}\left[F\left(\boldsymbol{\theta}\left(t+1\right)\right)\right]$ and $F^*$. A smaller upper bound would potentially narrow the gap. Particularly, $\mathbb{E}\left[||\boldsymbol{\theta}\left(t+1\right)-\boldsymbol{\theta}^*||_2^2\right]$ in \eqref{eq:opt_localSGD} can be interpreted as variance of the aggregated model produced from the model updates of the scheduled devices.
We list some system factors that affect it as follows.

\subsubsection{Training data distribution}
If all the devices have i.i.d. training data, we expect that $\Gamma\rightarrow0$, as discussed in Sec. \ref{sec:system_model}. It directly follows that
\begin{equation*}
	F^*\rightarrow\sum_{k\in\mathcal{N}}\frac{|\mathcal{S}_k|}{|\mathcal{S}|}F_k^*,
\end{equation*}
which implies a small $\mathbb{E}\left[||\boldsymbol{\theta}\left(t+1\right)-\boldsymbol{\theta}^*||_2^2\right]$ since $\boldsymbol{\theta}^*_k\simeq \boldsymbol{\theta}^*$, $\forall k\in\mathcal{N}$.
However, with non-i.i.d. training data, $\boldsymbol{\theta}^*_k$ may vary a lot across different devices, which leads to a large variance of the aggregated model.

\subsubsection{Data compression}
As the model updates are transmitted through rate-limited wireless channels, the server only receives noisy information due to data compression. Larger compression loss will give a larger variance of the aggregated model.

\subsubsection{Asynchronous model updates}
As shown in Sec. $\ref{sec:ayncFL}$, the received updates $\boldsymbol{u}_{k}(t), \forall k$ might have different ALUs, which will be an extra source of variation in $||\boldsymbol{\theta}\left(t+1\right)-\boldsymbol{\theta}^*||_2^2$.

From the perspective of device scheduling, we can achieve smaller $\mathbb{E}\left[||\boldsymbol{\theta}\left(t+1\right)-\boldsymbol{\theta}^*||_2^2\right]$ by considering the following aspects:
\begin{itemize}
	\item The scheduled devices should construct a homogeneous representation of the entire training data set $\mathcal{S}$.
	\item Compression loss needs to be kept low, which motivates us to prioritize devices with better channel conditions.
\end{itemize}
From the perspective of model aggregation, we can alleviate the adverse impact of asynchronous updates by considering age-aware weighting design in the aggregation process.
	
\section{Scheduling and Aggregation Design for Asynchronous FL}
We propose a scheduling policy that aims at achieving a smaller optimality gap, by considering the training data distribution and the channel conditions of ready-to-update devices.\footnote{Our scheduling design is applicable to any distributed ML setting with heterogeneous and unbalanced data, not only in the considered asynchronous FL setting.} Then, the aggregation weights are adjusted accordingly to alleviate the harmful impact from asynchronous training.

\subsection{Channel-aware Data-importance-based Scheduling}
To illustrate the idea, we consider a classification problem with labeled data. The training data set is represented by $\mathcal{S}=\{(x,y)|y\in\mathcal{L}\}$, where $\mathcal{L}=\{y_1,...,y_\omega\}$ is a finite set that contains all labels and $|\mathcal{L}|=\omega$. For any device $k$, $\boldsymbol{b}_k=[b_k^{1},...,b_k^{\omega}]$ is defined as the label distribution in $\mathcal{S}_k$, where $b_k^{j}$ is the number of $y_j$-labeled samples and $\sum_{j=1}^{\omega}b_k^{j}=|\mathcal{S}_k|$. To construct a homogeneous data distribution, we may select $\Pi(t)$ to achieve the minimal label variance
\begin{equation*}
	\Omega\left(\Pi(t)\right)=\sum_{j=1}^{\omega}\Big|\sum_{k\in\Pi(t)}b_k^{j}-\bar{b}\Big|^2,
\end{equation*}
where $\bar{b}=\frac{1}{\omega}\sum_{j=1}^{\omega}\sum_{k\in\Pi(t)}b_k^{j}$.
Additionally, scheduling devices with better channel quality leads to smaller compression loss per device. Combine these two selection criteria, we first pick a subset $\Pi'(t)\subseteq\mathcal{K}(t)$ comprising of devices with $\mbox{min}(0.5N, |\mathcal{K}(t)|)$ highest channel capacity $C_k(t)$. Next, we find $\Pi(t)
=\underset{\Pi\subseteq\Pi'(t)}{\mbox{argmin}} \,\Omega\left(\Pi\right)$.
	
\subsection{Age-aware Model Aggregation}
To tackle the asynchrony in model aggregation, we assign the weights not only based on the data proportion as in \eqref{eq:syncFlAggregation}, but also on its ALU.
The age-aware weighting design follows
\begin{equation}
	w_k(t)=\frac{|\mathcal{S}_k|{\gamma}^{a_k(t)}}{\sum_{i\in\Pi(t)}|\mathcal{S}_i|{\gamma}^{a_i(t)}}, \forall k\in\Pi(t).
	\label{eq:wk_t_age}
\end{equation}
Here, $\gamma$ is a real-valued constant, and the choice of its value can be divided into three cases:
\begin{itemize}
	\item $\gamma>1$, of which the system favors older local updates.
	\item $\gamma<1$, of which the system favors fresher local updates.
	\item $\gamma=1$, which is equivalent to the baseline design in \eqref{eq:syncFlAggregation}.
\end{itemize}
Favoring older updates could potentially balance the participation frequency among the devices and reduce the risk of model training biased towards those with stronger computing capability. This design performs well when data distribution is highly non-i.i.d. and some devices with inferior computing capability possess unique training data. However, it also creates the problem of applying outdated updates on the already-evolved model. On the other hand, favoring fresher local updates would help the model to converge fast and smoothly with time, at the risk of converging to an imbalanced model biased towards devices with superior computing power. 
Since the proposed scheduling policy is conducted in a way that improves the homogeneity of data distribution, the advantage of ``favoring fresher models'' strategy becomes more convincing. 

To summarize, the channel-aware data-importance-based scheduling design with favoring-fresh aggregation policy would lead to fast and smooth convergence performance. To verify the effectiveness of our design, we provide convergence analysis in Sec. \ref{sec:conv_analysis} and simulation results in \ref{sec:simulation_results}, respectively. 
	
\section{Convergence Analysis}
\label{sec:conv_analysis}
\begin{figure*}[t!]
	\centering
	\includegraphics[scale=.45]{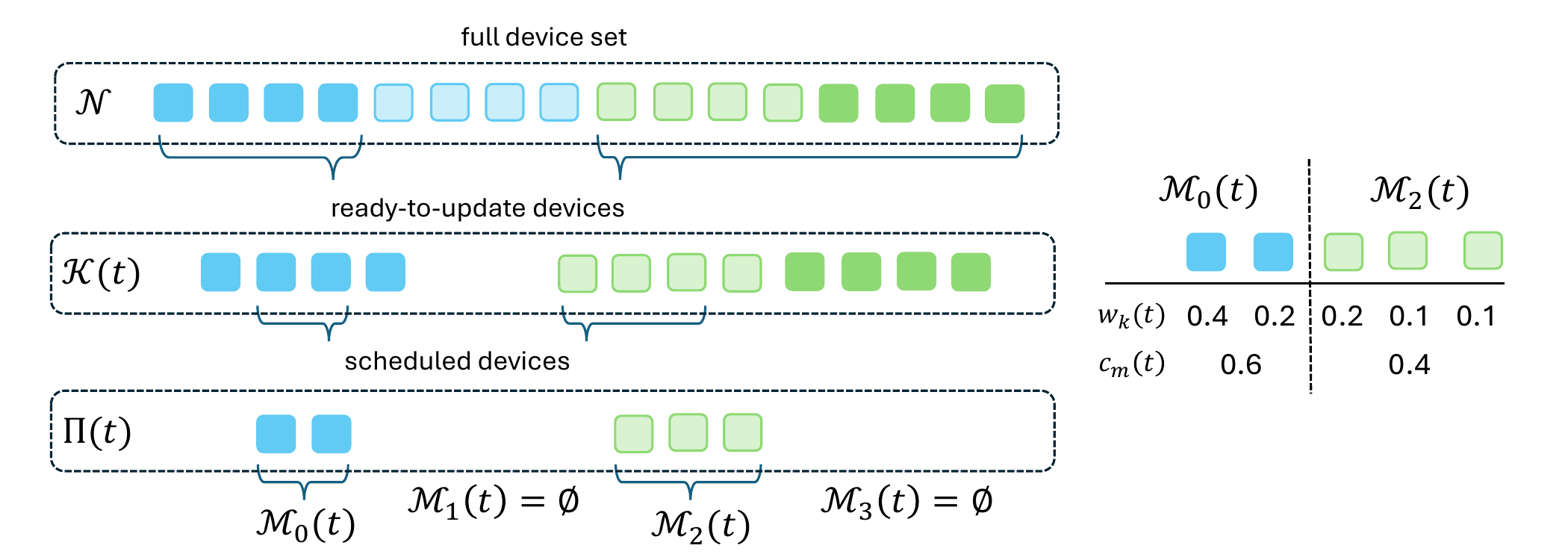}
	\caption{The relations between $\mathcal{N}$, $\mathcal{K}(t)$, $\Pi(t)$, and $\{\mathcal{M}_m(t)\}_0^{a_{\lim}}, a_{\lim}=3,$ at iteration $t$, where each colored block represents one device, and same-color devices have the same ALU. An example of normalization scaling for each nonempty $\mathcal{M}_m(t)$ is provided.}
	\label{fig:variableFig}
\end{figure*}

We introduce some notations and definitions for the convergence analysis of the proposed system.
\begin{definition}(\emph{Aggregation asynchrony}):	
	\begin{itemize}
		\item Let $a_{\lim}\triangleq\lceil T_{\max}/\tilde{T}\rceil$ be the maximum age of the received global model at a device.\footnote{Recall that $\tilde{T}$ is the aggregation period and $T_{\max}$ is the maximum device training time.} Then, there are at most $a_{\lim}+1$ different versions of received global model during the model aggregation, as $0\leq a_k(t)\leq a_{\lim}$.
		\item Let $\mathcal{M}_i(t)\subseteq\Pi(t), i=1,...,M(t)$ be a device subset with the same ALU, i.e., $\forall k,j\in\mathcal{M}_i(t)$, $a_k(t)=a_j(t)$.
	\end{itemize}
\end{definition}
The relations between the different quantities in the algorithm are illustrated in Fig. \ref{fig:variableFig}.
We introduce the following metric to quantify the data heterogeneity level for a specific device subset $\mathcal{M}$.
\begin{definition}(\emph{Data heterogeneity metric}): 	
	For a device subset $\mathcal{M}\subseteq\mathcal{N}$ and weights $\{w_k|k\in\mathcal{M}\}$ that satisfy $\sum_{k\in\mathcal{M}}w_k=1$, we define
	\begin{equation*}
		\Gamma_{\theta}(\mathcal{M})=\sum_{k\in\mathcal{M}}w_k\|\boldsymbol{\theta}^*-\boldsymbol{\theta}^*_k\|_2^2,
	\end{equation*}
	where $\boldsymbol{\theta}^*=\argmin F(\boldsymbol{\theta})$ and $\boldsymbol{\theta}_k^*=\argmin F_k(\boldsymbol{\theta})$.
	The weights $\{w_k|k\in\mathcal{M}\}$ are deterministically decided as functions of $ \mathcal{M}$ according to the local training data size.
	Also, define
	\begin{equation}
		\zeta=\underset{\mathcal{M}\subseteq\mathcal{N}}{\max}\Gamma_{\theta}(\mathcal{M}),
		\label{eq:noniid_b1}
	\end{equation}
	which is well-defined as  $\mathcal{N}$ is a finite set.
\end{definition}

We make the following assumptions on the local loss functions, stochastic gradient computation, and wireless link conditions:
\begin{assumption}
	(Smoothness): The local loss function $F_k(\boldsymbol{\theta}),\forall k$ is $L$-smooth, as specified in \eqref{asumpt:L-smooth}, 
	which is equivalent to Lipschitz continuous condition \cite{bottou2018ML}, i.e., $\forall \boldsymbol{\theta}_1,\boldsymbol{\theta}_2\in\mathbb{R}^d$,
	\begin{equation*}
		\|\nabla F_k\left(\boldsymbol{\theta}_1\right)-\nabla F_k\left(\boldsymbol{\theta}_2\right)\|_2\leq{L}\|\boldsymbol{\theta}_1-\boldsymbol{\theta}_2\|_2;
	\end{equation*}
	\label{asump:lSmooth}
\end{assumption}
\begin{assumption}
	(Strong convexity): The local loss function $F_k(\boldsymbol{\theta}),\forall k$ is $\mu$-strongly-convex, i.e., $\forall \boldsymbol{\theta}_1, \boldsymbol{\theta}_2\in\mathbb{R}^d$,
	\begin{equation*}
		F_k\left(\boldsymbol{\theta}_1\right)-F_k\left(\boldsymbol{\theta}_2\right)\geq\nabla F_k\left(\boldsymbol{\theta}_2\right)^T\left(\boldsymbol{\theta}_1-\boldsymbol{\theta}_2\right)+\frac{\mu}{2}\|\boldsymbol{\theta}_1-\boldsymbol{\theta}_2\|_2^2;
	\end{equation*}
\end{assumption}
\begin{assumption}
	(First and second moment constraints): $\forall \boldsymbol{\theta}\in\mathbb{R}^d$, the stochastic gradient $\nabla F_k(\boldsymbol{\theta};\mathcal{B}_k),\forall k$ satisfy
	\begin{align}
		\mathbb{E}_{\mathcal{B}_k}\left[\nabla F_k\left(\boldsymbol{\theta};\mathcal{B}_k\right)\right]&=\nabla F_k\left(\boldsymbol{\theta}\right),\nonumber\\
		\mathbb{E}_{\mathcal{B}_k}\left[\|\nabla F_k\left(\boldsymbol{\theta};\mathcal{B}_k\right)\|_2^2\right]&\leq C_1+C_2\|\nabla F_k\left(\boldsymbol{\theta}\right)\|_2^2,
		\label{eq:asmp3_2}
	\end{align}
	for some positive constants $C_1$ and $C_2$.
	\label{asump:sgdCt}
\end{assumption}
\noindent Note that it appears to be common in the literature to assume uniformly
bounded gradient, i.e., $C_2=0$ in \eqref{eq:asmp3_2}, and a strongly convex function over an unbounded space, e.g. \cite{chen2020asynchronous,wang2021resource,amiriconvergence,salehi2021federated,chen2020aJoint,chai2020fedat,li2019convergence,stich2018local}.
These assumptions, however, are fundamentally incompatible as there exists no strongly convex function
whose gradient is uniformly bounded over $\mathbb{R}^d$, which is also pointed out in \cite{Lam2019NewConv}. Working with a strongly convex function
and uniformly bounded gradients requires a bounded parameter space, which yields a different (constrained) optimization problem, for which a solution algorithm would have to incorporate a projection step. We thus assume
that the ``noise'' in the gradients is uniformly bounded (in expectation), also adopted in \cite{bottou2018ML}, which allows us to work with a strongly convex objective over an unbounded space.
\begin{assumption}
	There exists an $r_{\min}>0$ such that $\forall k\in\Pi(t)$ and for all $t$,
	\begin{equation}
		\mathbb{E}[r_k(t)]\geq r_{\min},
		\label{ieq:rmin}
	\end{equation}
	where $r_k(t)$ is the number of preserved elements in the sparsified model update vector. The expectation is with respect to the randomness in the wireless channels. The intuition behind the condition $r_{\min}>0$ is that with our channel-aware device scheduling policy, the scheduled devices should have good channel conditions and thus require less  sparsification. 
	\label{assump:rmin}
\end{assumption}

In the following theorem, we provide the main convergence result of our system model in the special case with one local iteration per communication round, i.e., $E=1$.
For simplicity, for the mini-batch $\mathcal{B}_k(t,0)$, we omit the second argument and write
$\mathcal{B}_k(t)$. Moreover, the regularized local loss function $H_k\left(t,0\right)$ degenerates to $F_k\left(\boldsymbol{\theta}_{k}(t,0)\right)$, and thus
\begin{equation*}
	\nabla H_k\left(t,0\right)=\nabla F_k\left(\boldsymbol{\theta}_{k}(t,0);\mathcal{B}_k(t)\right).
\end{equation*}
\begin{theorem}
	Under Assumptions \ref{asump:lSmooth}-\ref{assump:rmin}, $E=1$, constant learning rate \mbox{$\alpha(t)\triangleq\alpha<\frac{\mu r_{\min}}{d\left(2L^2+C_3\right)}$},
	and partial device participation such that \mbox{$\Pi(\rho)=\cup_{j=0,...,a_{\lim}}\mathcal{M}_j(\rho)\subseteq\mathcal{N}, \rho=1,...,t$}, it holds that
	\begin{align}
		&\mathbb{E}\left[\|\boldsymbol{\theta}\left(t+1\right)-\boldsymbol{\theta}^*\|_2^2\right]\nonumber\\
		&\leq\left[1-\mu\alpha r_{\min}/d+\alpha^2\left(2L^2+C_3\right)\right]^{\lfloor\frac{t-1}{a_{\lim}+1}\rfloor+1}\mathbb{E}\left[\|\boldsymbol{\theta}(1)-\boldsymbol{\theta}^*\|_2^2\right]\nonumber\\
		&\quad+\frac{\epsilon}{\mu r_{\min}/d-\alpha\left(2L^2+C_3\right)},\label{ieq:thm_rslt}
	\end{align}
	where $C_3=8L^2\left[\left(1+\frac{d}{4\nu^2}\right)C_2+1\right]$ and
	\begin{equation}
		\epsilon=
		\zeta L+\alpha\left[\zeta\left(2L^2+C_3\right)+4C_1\left(1+\frac{d}{4\nu^2}\right)\right].\label{eq:epsilon}
	\end{equation}
	The expectation is taken over the randomness of the stochastic gradients, the channel gains and the model compression, and the device scheduling during all the past iterations.
\end{theorem}
\noindent\emph{Proof.} See Section \ref{proofThm1}. \hfill$\qed$

\vspace{.3cm}
\noindent Note that because of the $L$-smoothness (Assumption \ref{asump:lSmooth}), the optimality gap can be bounded as $$\mathbb{E}\left[F(\boldsymbol{\theta}(t+1))\right]-F^* \le\frac{L}{2}\mathbb{E}\left[\|\boldsymbol{\theta}\left(t+1\right)-\boldsymbol{\theta}^*\|_2^2\right].$$
\begin{remark}
	The impact of multiple local iterations, i.e., $E>1$, on the convergence analysis has been already established and exploited in existing literature on FL (see, e.g., \cite{stich2018local},\cite{singh2021comm,koloskova2020unified}). 
	The focus of our analysis is the impact of update compression, training data heterogeneity, and intra-iteration asynchrony on the system convergence. 
	We believe that extending our results to the case with $E>1$ is possible by following approaches from, for example, 
		\cite{stich2018local}, \cite{singh2021comm} or \cite{koloskova2020unified};
		this, however, would require, among other things, quantifying local model divergence.\footnote{An extra term $||\boldsymbol{\theta}_k(t,\tau)-\boldsymbol{\theta}_k(t,0)||_2^2$ will be introduced and could be bounded by its gradient $||\nabla F_k(\boldsymbol{\theta}_k(t,\tau))||_2^2$ through \eqref{eq:asmp3_2}. The learning rate would then have to be redesigned accordingly.}
	For analytical clarity and simplicity, we consider the case with $E=1$.	
\end{remark}
\begin{remark}
	\label{rmk:conv_asymp}
	Based on \textbf{Theorem 1}, the upper bound on the error asymptotically converges to the following constant:
	\begin{equation*}
		\frac{\zeta L+\alpha\left[\zeta\left(2L^2+C_3\right)+4C_1\left(1+\frac{d}{4\nu^2}\right)\right]}{\mu r_{\min}/d-\alpha\left(2L^2+ C_3\right)}
	\end{equation*}
	when $t\rightarrow\infty$. 
	This asymptotic error  increases with $\zeta$ and decreases with increasing $r_{\min}$ and $\nu$. 
	This  reflects that better learning performance can be obtained by
	\begin{itemize}
		\item a lower level of data heterogeneity (i.e.,  smaller $\zeta$),
		\item scheduling devices with better channel quality, resulting in a larger $r_{\min}$,
		\item and a higher number of quantization levels (i.e.,  higher $\nu$),
		\item a smaller learning rate $\alpha$.
	\end{itemize}
	Note that this asymptotic error bound can be  small but does not vanish due to the 
	effects of partial device participation and data heterogeneity. This is a consequence of the scheduling algorithm.
\end{remark}
\begin{remark}
	\label{rmk:conv_perIter}
	A better link quality (higher $r_{\min}$) and higher-resolution updates (higher $\nu$, leading to lower $C_3$) leads to faster convergence towards the asymptotic error, as indicated in \eqref{ieq:thm_rslt}.
	In the analysis, the adverse impact of the update asynchrony on the learning performance is reflected by the extra terms $\|\boldsymbol{\theta}(t-m)-\boldsymbol{\theta}^*\|_2^2, m=0,...,a_{\lim}$. In \eqref{ieq:thm_rslt}, this results
	in a slowing down of the  contraction speed from $(1-\mu\alpha)^{2t}$  nominally for vanilla gradient descent \cite[Th. 3.12]{bookNew}
	on a smooth, strongly convex function, 
	to $\left[1-\mu\alpha r_{\min}/d+\alpha^2\left(2L^2+C_3\right)\right]^{\lfloor\frac{t-1}{a_{\lim}+1}\rfloor+1}$. 
	
\end{remark}
These two remarks confirm the importance of considering both training data distribution and channel quality in our device scheduling design.

\section{Simulation Results}
\label{sec:simulation_results}
We perform simulations using the MNIST \cite{lecun-mnisthandwrittendigit-2010} data set for solving the hand-written digit classification problem by adopting a convolutional neural network with model dimension $d=21840$. The block fading channel spans $n$ symbols for uplink transmissions. We consider Rayleigh fading $h_k(t)\sim \mathcal{CN}(0,1)$ and uplink power control such that the received signal-to-noise ratio is $13$dB.
The system parameters are specified as follows.
\begin{itemize}
	\item (Training data distribution) $|\mathcal{S}|=60000$ samples are distributed evenly to all the devices. Both i.i.d. and non-i.i.d. data distribution scenarios are considered. For the i.i.d. case, $|\mathcal{S}|/N$ samples are randomly allocated to each device without replacement. For the non-i.i.d. case, the data allocation follows the setup in \cite{mcmahan2017communication}, where each device contains up to $\mbox{min}(\lfloor200/N\rfloor,10)$ different digits. 
	\item (Heterogeneous computing capability) We use $T_k$ as the local training duration of device $k$, generated from uniform distribution $T_k\sim\mathcal{U}(T_{\min},T_{\max})$, where $T_{\min}$ is the least possible device training time.
	\item (Compression) $4$-level random quantizer, $\mathcal{Q}_4$, is adopted.
	\item (Diminishing learning rate) $\alpha(1,0)=0.01$ is initially adopted, together with regularization coefficient $\lambda=0.02$.
\end{itemize}

\begin{figure}[t!]
	\centering
	\includegraphics[width=.8\columnwidth]{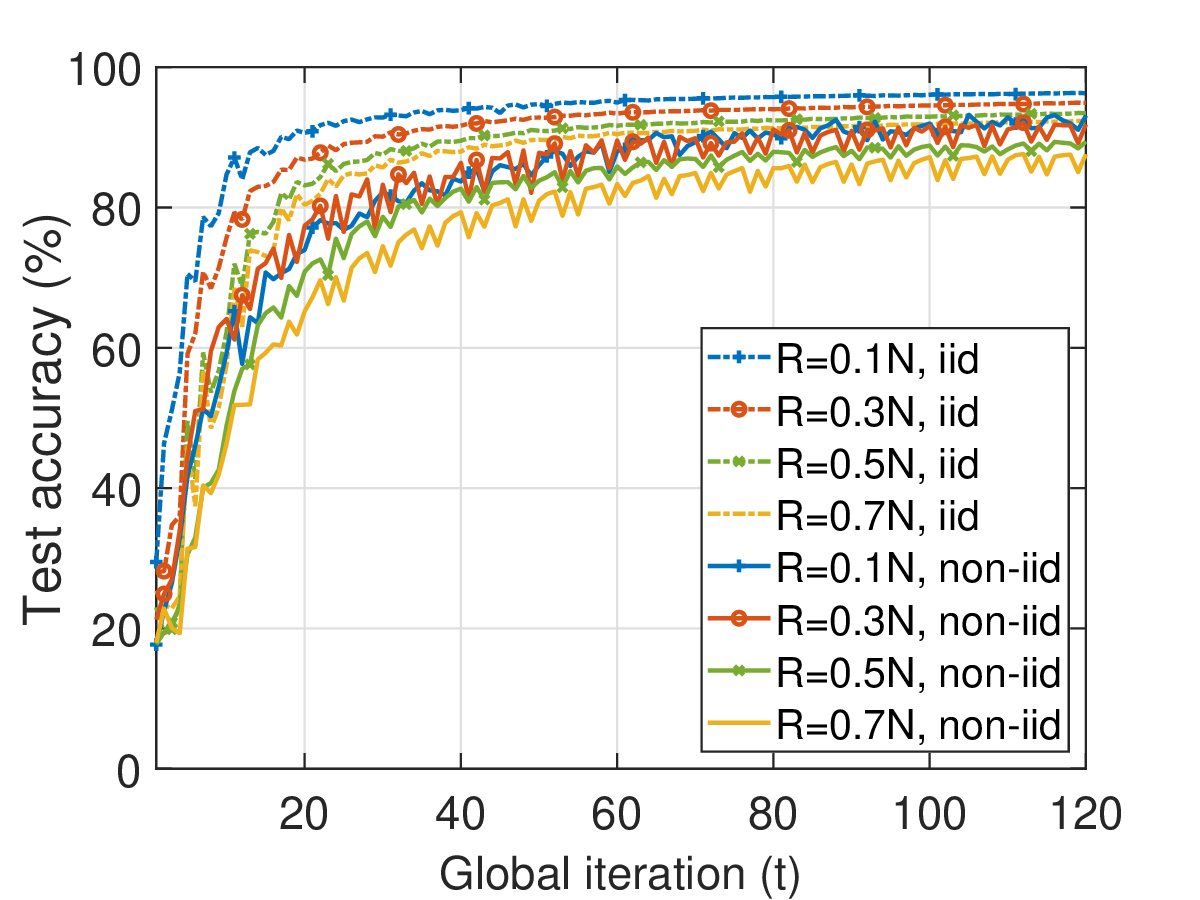}
	\caption{Test accuracy of the proposed scheme under different partial scheduling ratio, where $N=40$ and $n=50000$.}
	\label{fig:diffR}
	\hfill
\end{figure}

Since the communication resources are shared by maximally $R$ devices, more scheduled devices means less communication resources per device, leading to a higher compression loss. 
The test accuracy results with various choices of $R$ are shown in Fig.~\ref{fig:diffR}. As we can see, in the i.i.d. scenario, a smaller $R$ is preferable since it gives received model updates with better precision as the result of more allocated bits per user. On the other hand, in the non-i.i.d. scenario, there exists a trade-off between  compression loss and  model bias, which makes the choice of $R$ important. In this work, we focus on the impact of the scheduling design for a fixed $R$. The optimal value of $R$ depends on many system parameters; its optimization could be studied in future work.

Finding the optimal value of $\tilde{T}$ analytically is highly non-trivial as we are short of a tractable  expression that  characterizes the relation between the optimality gap and the value of $\tilde{T}$. 
For the numerical results reported herein, we choose $\tilde{T}=T_\text{max}/4$ as this gives the best results in the simulations (see below for the  comparisons between different $\tilde{T}$).
\subsection{Comparison of FL frameworks}
\begin{figure}[t!]
	\centering
	\begin{subfigure}[b]{0.49\columnwidth}
		\centering
		\includegraphics[width=\columnwidth]{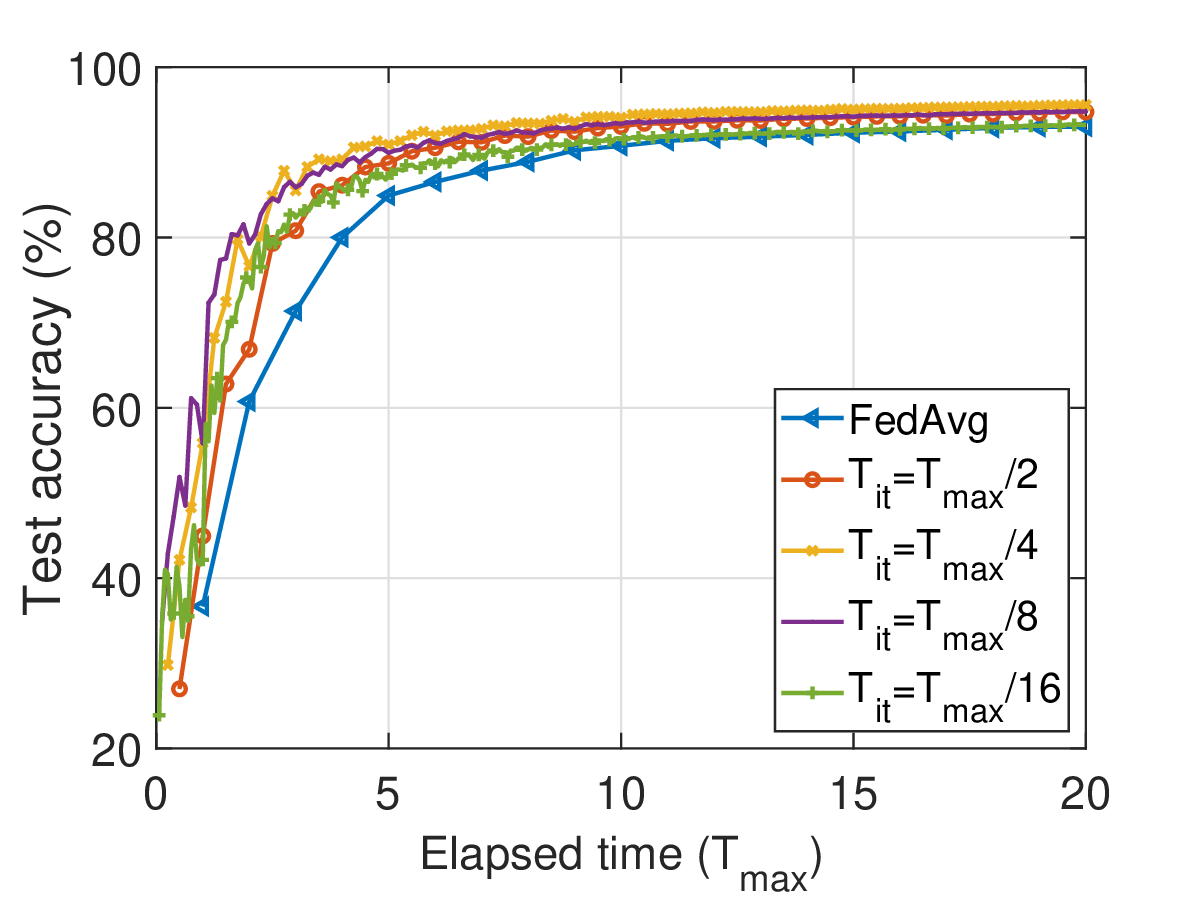}
		\caption{i.i.d. data}
		\label{fig:diffT_iid}
	\end{subfigure}
	\begin{subfigure}[b]{0.49\columnwidth}
		\centering
		\includegraphics[width=\columnwidth]{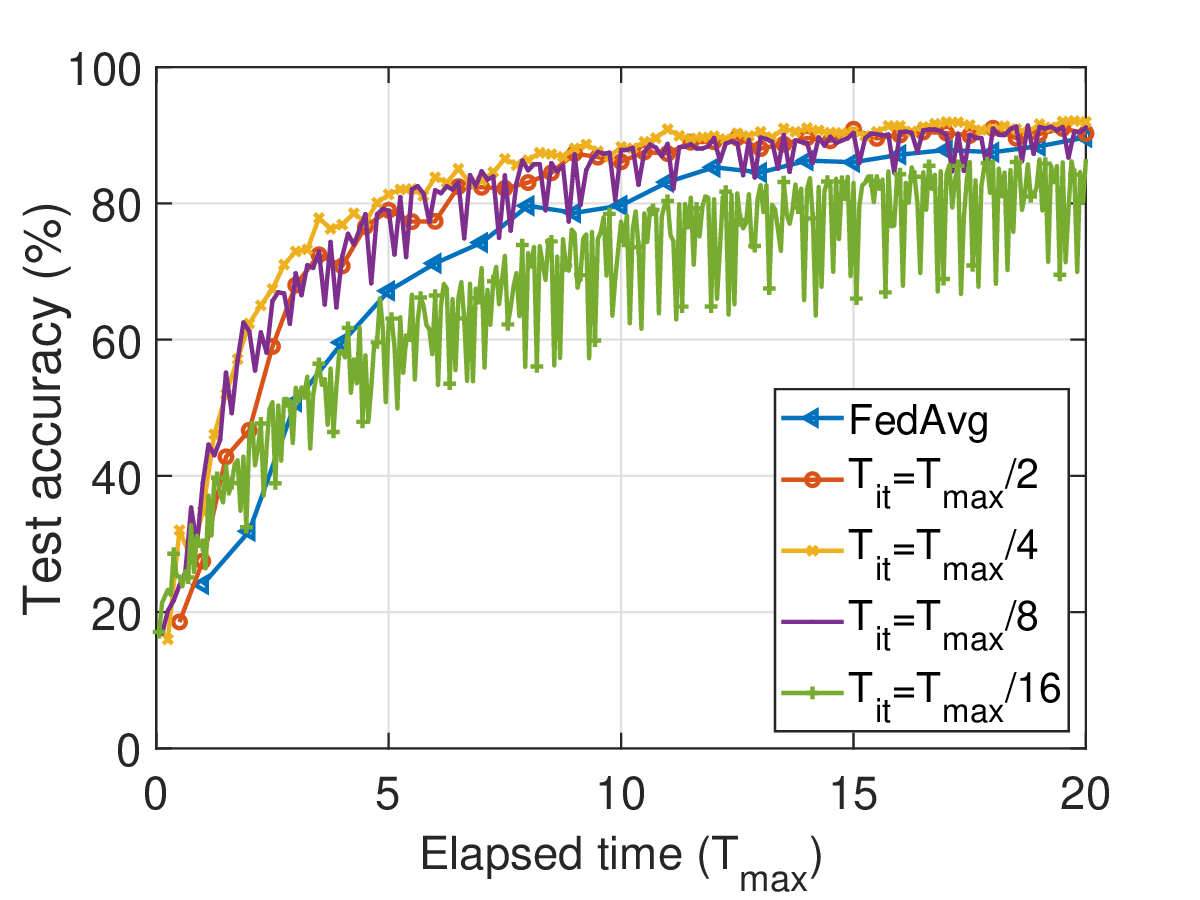}
		\caption{non-i.i.d. data}
		\label{fig:diffT_niid}
	\end{subfigure}
	\caption{Impact of $\tilde{T}$ on test accuracy of the proposed asynchronous FL with periodic aggregation in i.i.d. and non-i.i.d. scenarios, where $N=40$, $R=0.2N$ and $n=300000$.}
	\label{fig:diffT}
	\hfill
\end{figure}

\begin{figure}[t!]
	\centering
	\begin{subfigure}[b]{0.49\columnwidth}
		\centering
		\includegraphics[width=\columnwidth]{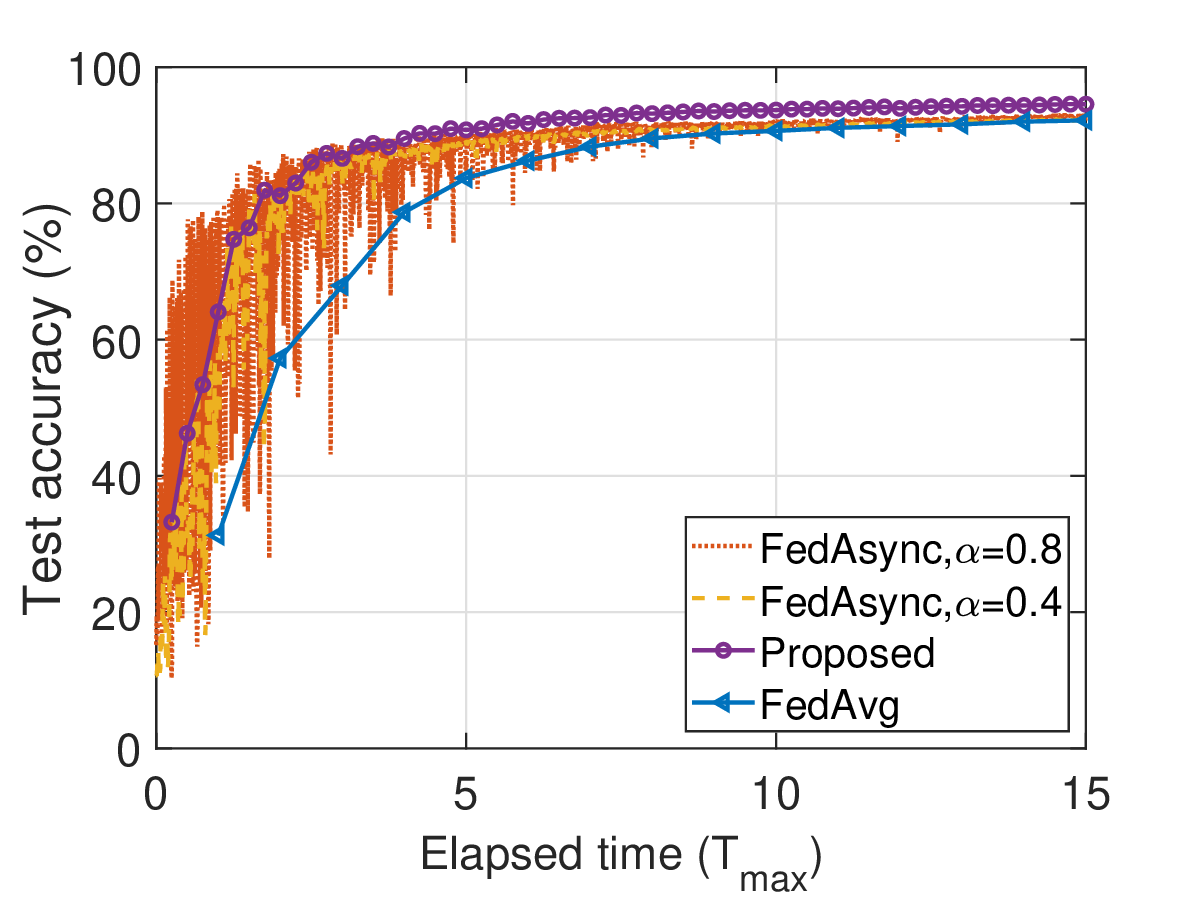}
		\caption{i.i.d. data}
		\label{fig:diffFed_iid}
	\end{subfigure}
	\begin{subfigure}[b]{0.49\columnwidth}
		\centering
		\includegraphics[width=\columnwidth]{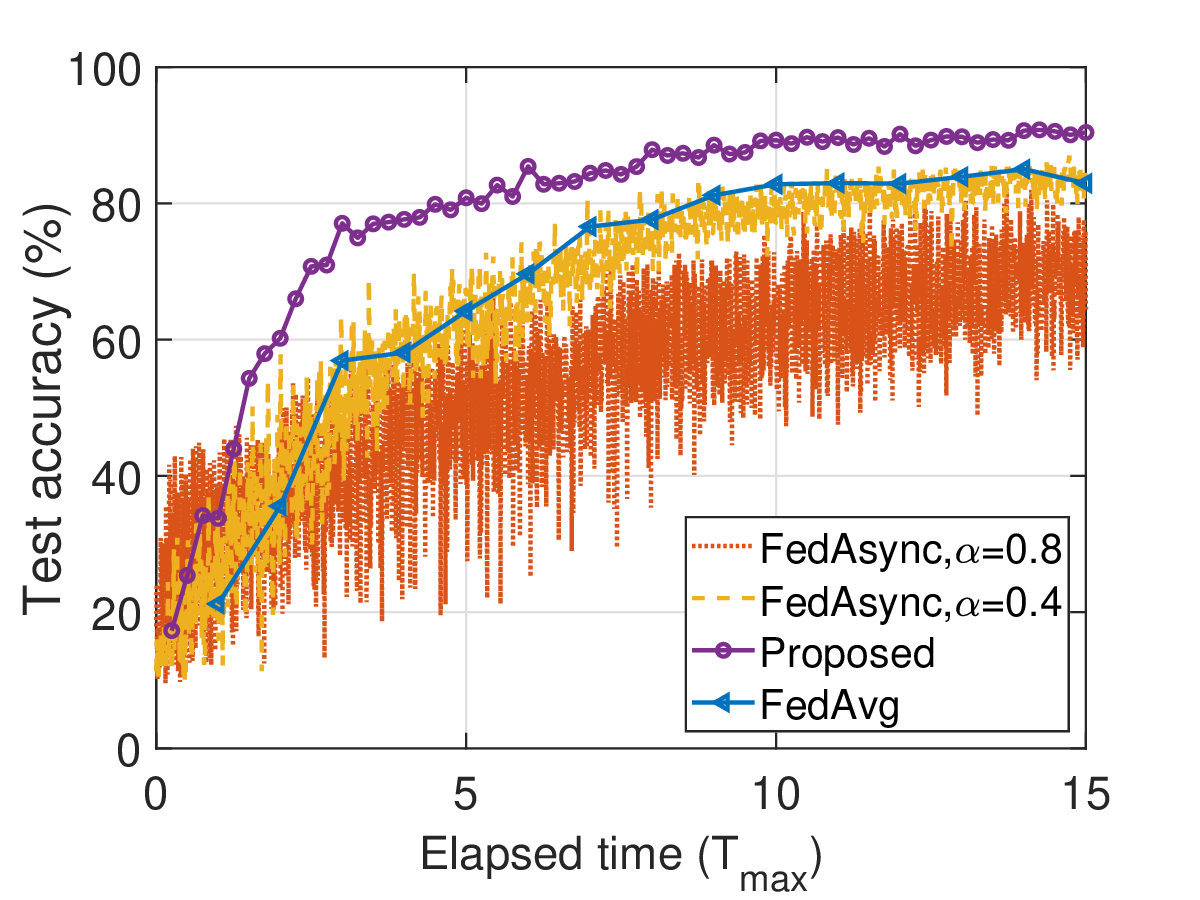}
		\caption{non-i.i.d. data}
		\label{fig:diffFed_niid}
	\end{subfigure}
	\caption{Test accuracy of FedAvg, the proposed asynchronous FL with periodic aggregation, and FedAsync \cite{xie2019asynchronous} in i.i.d. and non-i.i.d. scenarios, where $N=40$, $R=0.2N$ and $n=300000$.}
	\label{fig:diffFed}
	\hfill
\end{figure}

	First, in Fig. \ref{fig:diffT}, we present the performance of our proposed design (with random scheduling and baseline aggregation, i.e., $\gamma=1$ in \eqref{eq:wk_t_age}) obtained with different values of $T_\text{max}/\tilde{T}$, to demonstrate the effect of the aggregation time period on the convergence performance. The performance of synchronous FedAvg is also presented for comparison. Then, in Fig. \ref{fig:diffFed}, we show the performance comparison between FedAvg, our proposed asynchronous FL, and FedAsync \cite{xie2019asynchronous}.
	Note that to make a fair comparison between different schemes, we fix the average amount of communication resources allocated over a certain time period $T'$. For example, if we assume there are $n$ symbols per communication round when $\tilde{T}=T_{\max}/4$ is adopted, then $4n$ symbols per round will be available for FedAvg. If FedAsync executes $N_{\text{FA}}$ iterations over time duration $T'$, the available number of symbols  per round would be $4nT'/(T_{\max}N_{\text{FA}})$.

As shown in Fig. \ref{fig:diffT}, $\tilde{T}=T_{\max}/4$ gives the best performance. 
However, this conclusion cannot be generalized since the result depends on many other system parameters.

The comparison between our proposed design, FedAvg, and FedAsync\footnote{An $\alpha$-filtering mechanism is applied on device updates, specifically, $\boldsymbol{\theta}(t+1)=(1-\alpha)\boldsymbol{\theta}(t)+\alpha\boldsymbol{u}_k(t)$.} (with $\alpha=0.4$ and $\alpha=0.8$) is shown in Fig. \ref{fig:diffFed}. We observe that although FedAsync can help reduce the straggler effect, the test accuracy result shows strong fluctuation, especially with non-i.i.d. training data. For both i.i.d. and non-i.i.d. scenarios, our proposed asynchronous FL design outperforms FedAsync and FedAvg, which shows its effectiveness in eliminating the straggler effect and achieving better convergence performance.

\subsection{Validation of convergence analysis}
\begin{figure}[t!]
	\centering
	\includegraphics[width=.8\columnwidth]{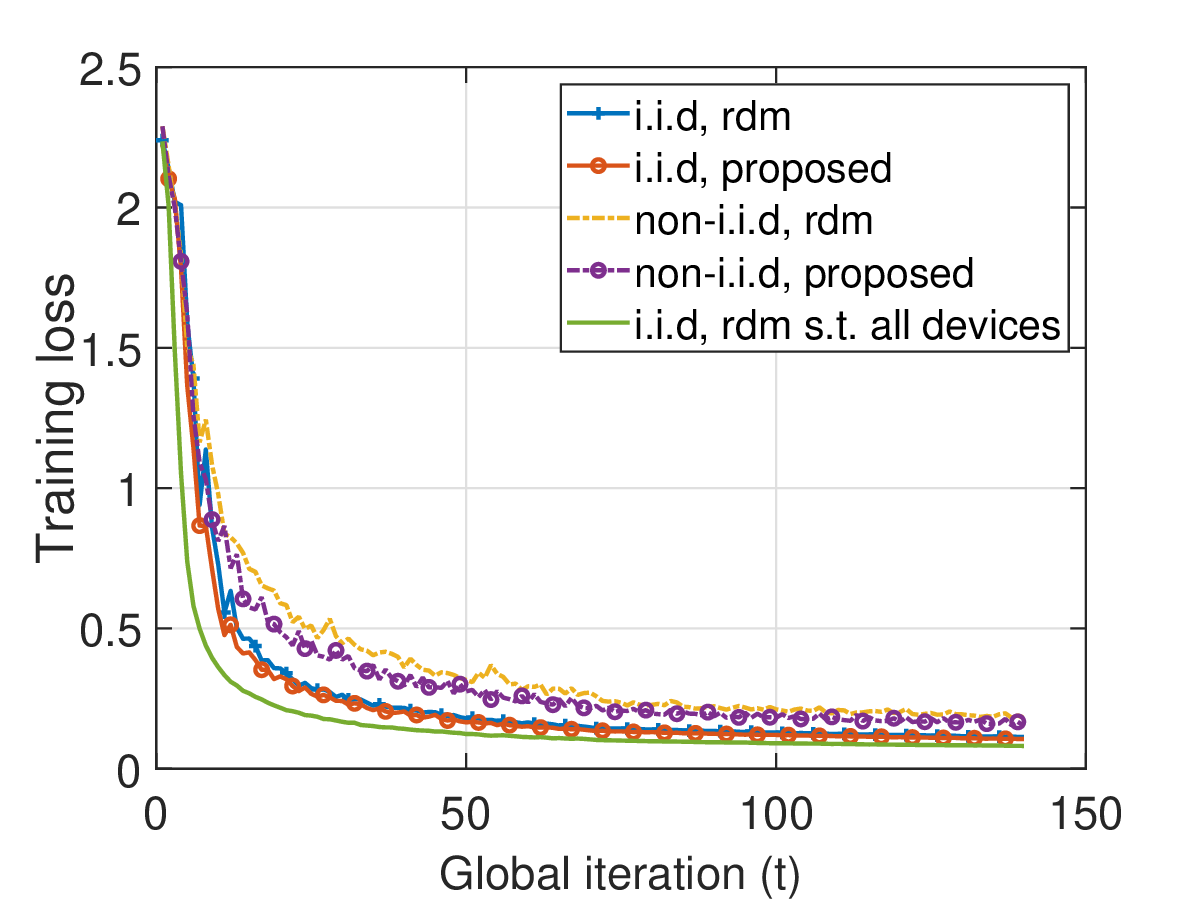}
	\caption{Training loss for random and the proposed scheduling methods in i.i.d. and non-i.i.d. scenarios, where $N=40$, $R=0.2N$ and $n=300000$.}
	\label{fig:trainLoss}
	\hfill
\end{figure}

We provide training loss comparison in Figure \ref{fig:trainLoss} to validate the conclusions drawn in Remarks \ref{rmk:conv_asymp} and \ref{rmk:conv_perIter}. We observe the following:
	\begin{itemize}
		\item With the random and proposed scheduling methods, the training loss is lower in the i.i.d. scenario than in the non-i.i.d. scenario, which shows that a smaller $A$ leads to a lower training loss.
		\item For both the i.i.d. and non-i.i.d. scenarios, our proposed scheduling design outperforms random scheduling, which validates the advantage of selecting devices with better link qualities and, collectively, a more homogeneous data representation.
		\item To show the impact of intra-iteration asynchrony, we present the result for synchronous updates with random scheduling for i.i.d. data (the green curve). As compared to the asynchronous case (blue curve), the synchronous case has lower training loss, which shows that intra-iteration asynchrony is harmful for the convergence performance.   
\end{itemize}

Moreover, as shown in Fig. \ref{fig:diffT_niid}, we see that with smaller $\tilde{T}$, which implies a larger $a_{\lim}$, the curves for the test accuracy result are less smooth than the ones with larger $\tilde{T}$. This validates the insights in Remark \ref{rmk:conv_perIter} that a higher degree of intra-iteration asynchrony leads to further degradation of the learning performance.

\subsection{Comparison of scheduling policies}
\begin{figure}[t!]
	\centering
	\begin{subfigure}[b]{0.9\columnwidth}
		\centering
		\includegraphics[width=\columnwidth]{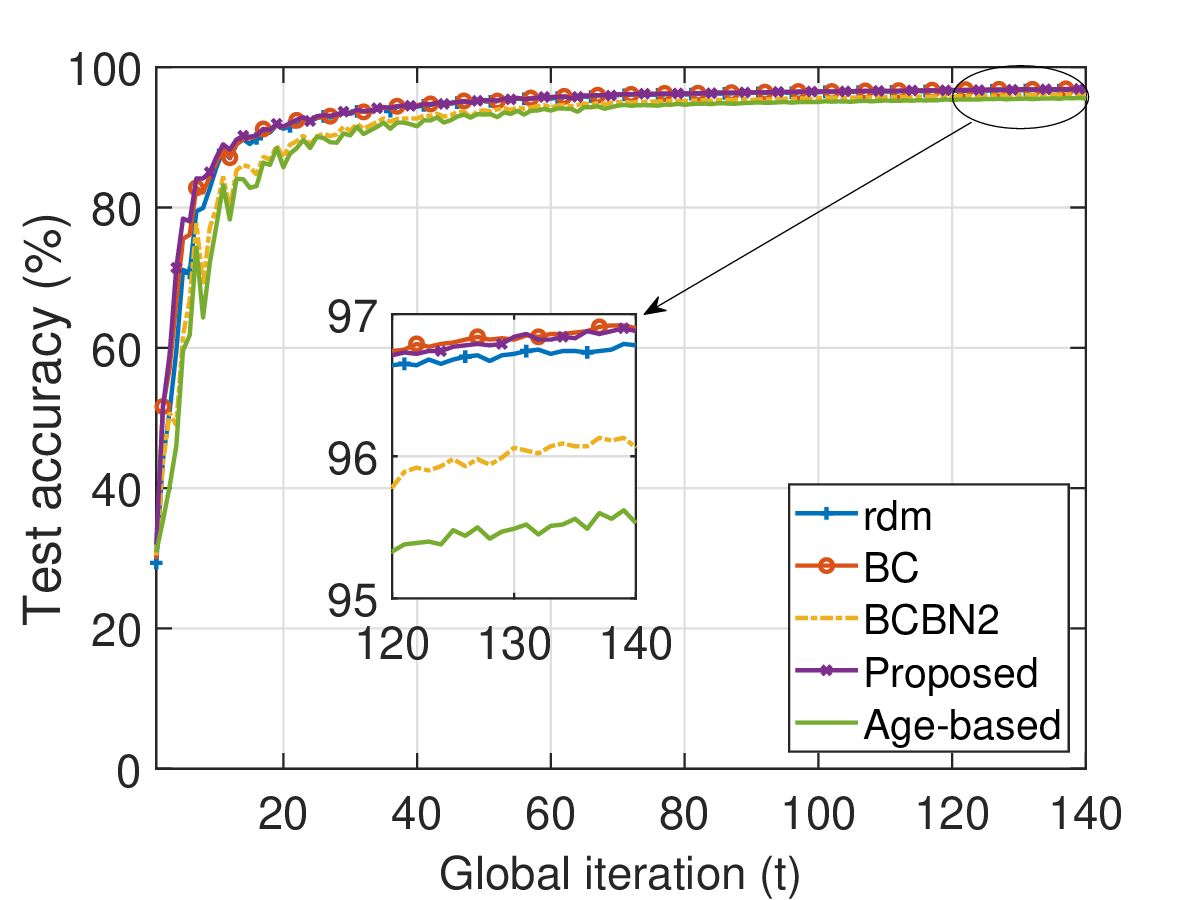}
		\caption{i.i.d. data}
		\label{fig:diffSch_iid}
	\end{subfigure}
	\begin{subfigure}[b]{0.9\columnwidth}
		\centering
		\includegraphics[width=\columnwidth]{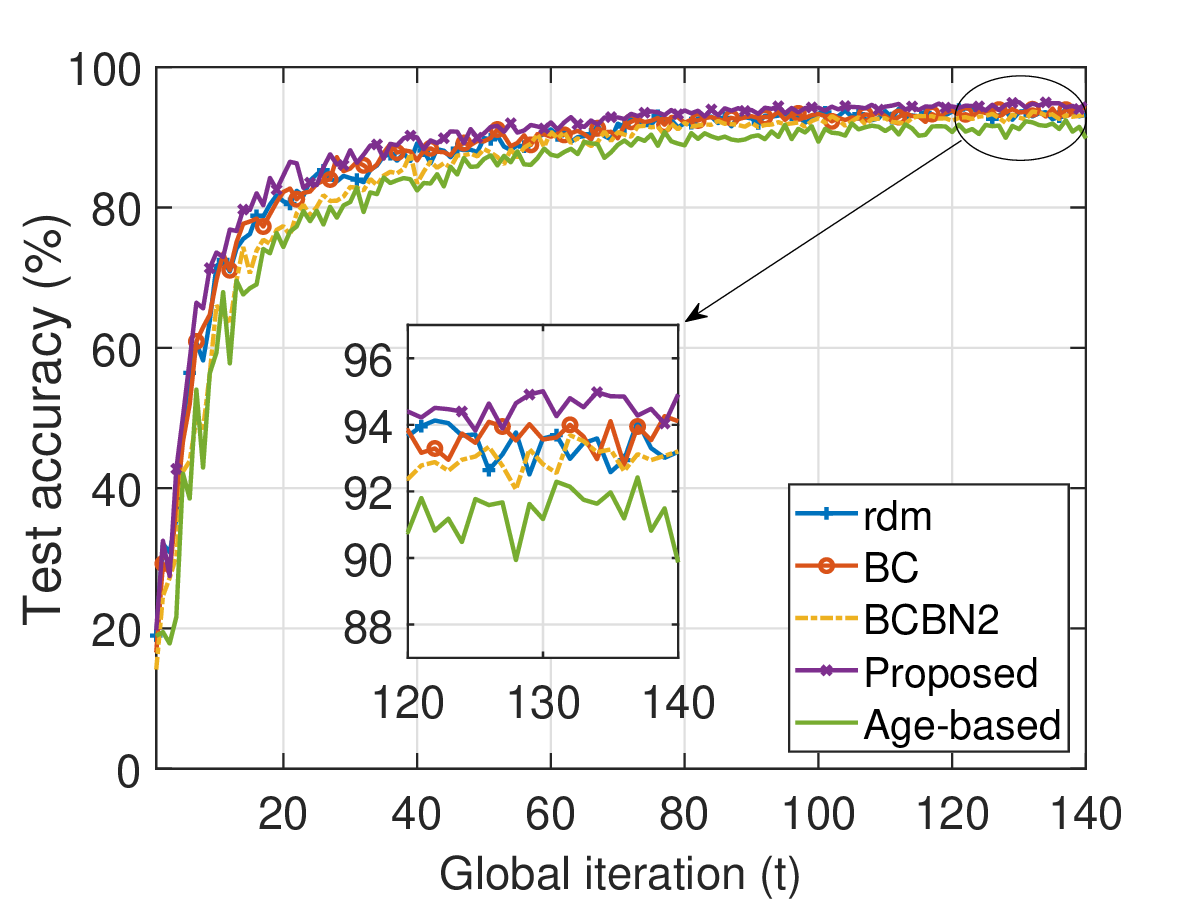}
		\caption{non-i.i.d. data}
		\label{fig:diffSch_niid}
	\end{subfigure}
	\caption{Test accuracy of different scheduling policies, where $N=40$, $R=0.2N$ and  $n=300000$.}
	\label{fig:diffSch}
	\hfill
\end{figure}

First, the aggregation weights are set with $\gamma=1$ in \eqref{eq:wk_t_age}. In Fig. \ref{fig:diffSch}, we show the test accuracy of the proposed scheduling design with some alternative reference methods.  
\begin{itemize}
	\item rdm: we select up to $R$ devices uniformly at random.
	\item BC \cite{amiriconvergence}: We select up to $R$ devices with highest $C_k(t)$.
	\item BCBN2 \cite{amiriconvergence}\footnote{For a fair comparison, symbol resource allocation follows \eqref{eq:nc}, differing from the design in \cite{amiriconvergence}.}: We first find a device subset $\mathcal{C}\subseteq\mathcal{K}(t)$ with $|\mathcal{C}|$ up to $0.5N$ that has highest $C_k(t)$. Then, among $\mathcal{C}$ we schedule devices with up to $R$ highest
	$||\boldsymbol{u}_k(t)||_2^2$.
    \item Age-based \cite{yang2019agebased}\footnote{This is a tweaked version of the method in \cite{yang2019agebased}, which likewise finds devices with better channel quality and minimizes the overall device staleness.}: We first find $\mathcal{C}$ using the same way as in BCBN2, then among $\mathcal{C}$ we schedule up to $R$ devices with highest staleness metric
    \begin{equation*}
	    c_k(t)=
	    \begin{cases}
		    0, & t=0 \\
		    \sum_{t'=0}^{t-1}\boldsymbol{1}(k\notin\Pi(t')), & \text{otherwise}
	    \end{cases}
    \end{equation*}
\end{itemize}
We see that the proposed scheduling policy outperforms the baseline random scheduling and the reference methods in both i.i.d. and non-i.i.d. scenarios. Besides, in non-i.i.d. scenario, we observe that the data-awareness in the proposed method further achieves a higher test accuracy than the pure channel-aware methods BC and BCBN2 \cite{amiriconvergence}.  

\subsection{Comparison of aggregation policies}
\begin{figure}[ht!]
	\centering
	\begin{subfigure}[b]{0.9\columnwidth}
		\centering
		\includegraphics[width=\columnwidth]{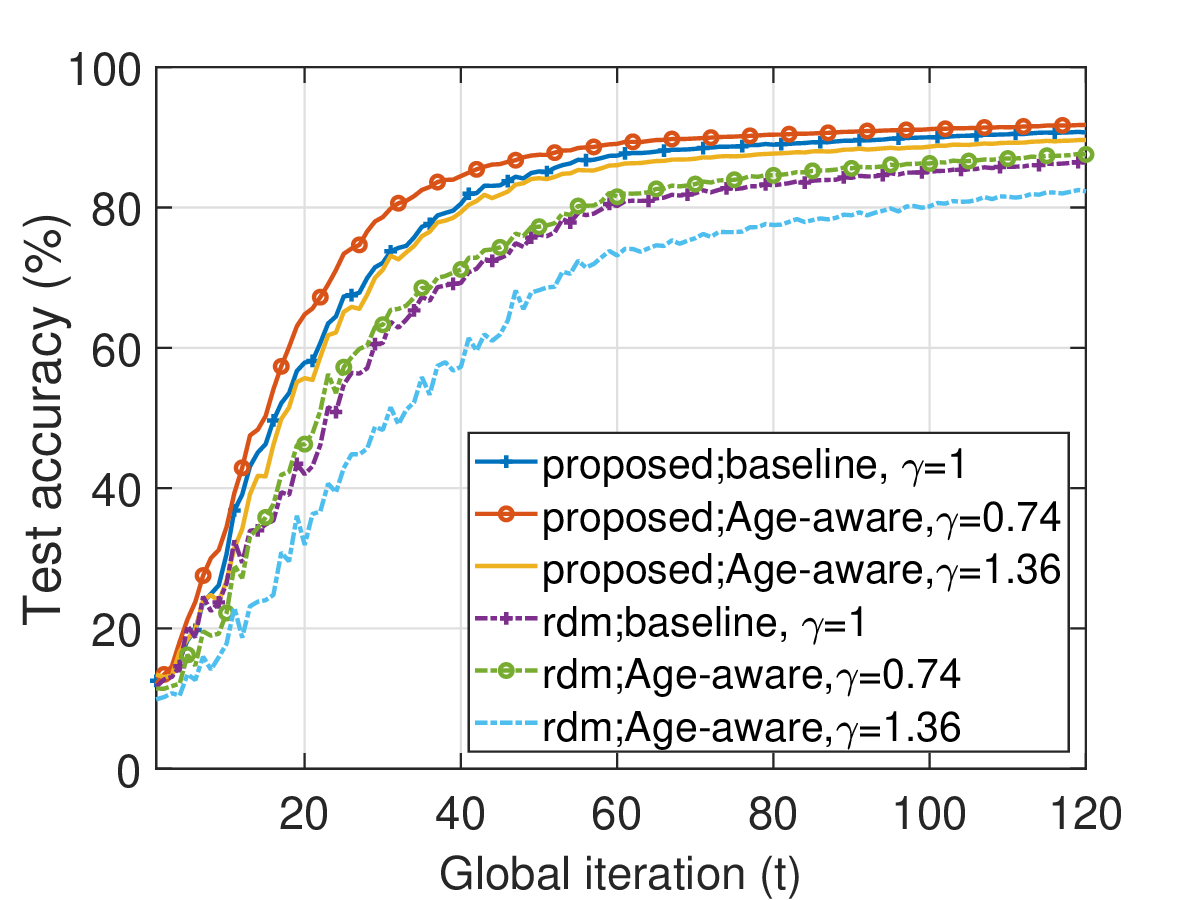}
		\caption{i.i.d. data}
		\label{fig:diffAggr_iid}
	\end{subfigure}
	\begin{subfigure}[b]{0.9\columnwidth}
		\centering
		\includegraphics[width=\columnwidth]{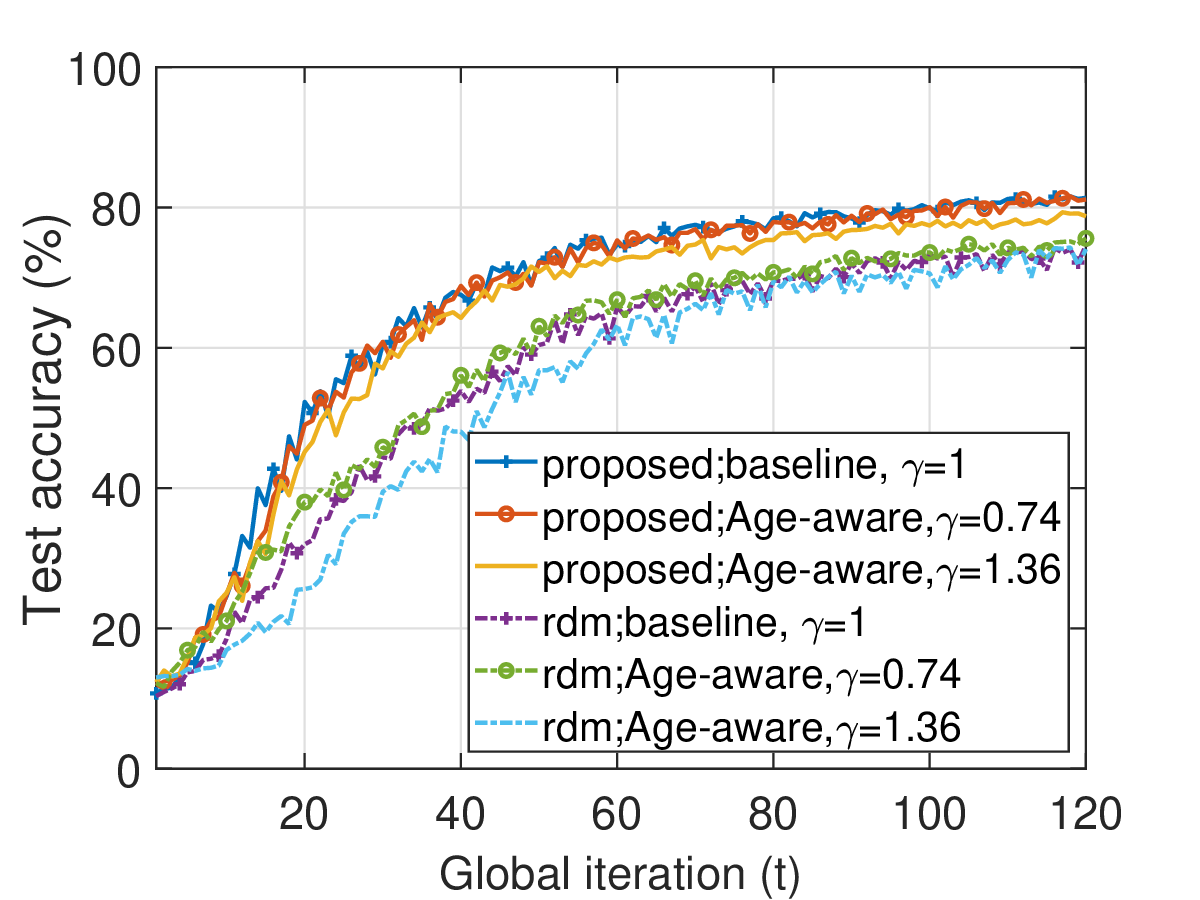}
		\caption{non-i.i.d. data}
		\label{fig:diffAggr_niid}
	\end{subfigure}
	\caption{Test accuracy of proposed aggregation policies, where $N=100$, $R=0.3N$ and $n=380000$.}
	\label{fig:diffAggr}
	\hfill
\end{figure}

In Fig.~\ref{fig:diffAggr}, we show the performance of different aggregation policies under the proposed channel-aware data-importance-based scheduling and random scheduling methods. We see that the age-aware design outperforms the baseline with $\gamma=1$ in both i.i.d. and non-i.i.d. scenarios. For i.i.d. case, the favoring-fresh weighting strategy has an outstanding performance gain. For non-i.i.d. case, performance gain of the proposed aggregation design is more obvious when random scheduling is adopted.

\section{Conclusions}
In this work, we proposed an asynchronous FL framework with periodic aggregation that combines the advantages of asynchronous training and synchronous model aggregation. For the proposed design, we further developed a channel-aware data-importance-based scheduling policy and age-based aggregation design for FL under wireless resource constraints. Our proposed scheduling and aggregation design was shown to outperform existing methods, especially with heterogeneous training data among different devices. The main takeaway message is that the design principle of scheduling and resource allocation in wireless FL should be based on reducing the bias and variance of aggregated local updates. For asynchronous FL settings, balancing the freshness and usefulness of local model updates and adjusting their contributions in the new global model is also an important design aspect. 

\section{Acknowledgement}
We thank Fredrik Jansson for his contribution to the idea of scheduling policy during his Master thesis project at the Division of Communication Systems, Linköping University.
	
\section{Appendix}
In the following, we simplify the notation of Euclidean norm $\|\cdot\|_2$ as $\|\cdot\|$.
\subsection{Proof of Theorem 1}
\label{proofThm1}

\begin{figure*}[t!]
	\centering
	\includegraphics[scale=.48]{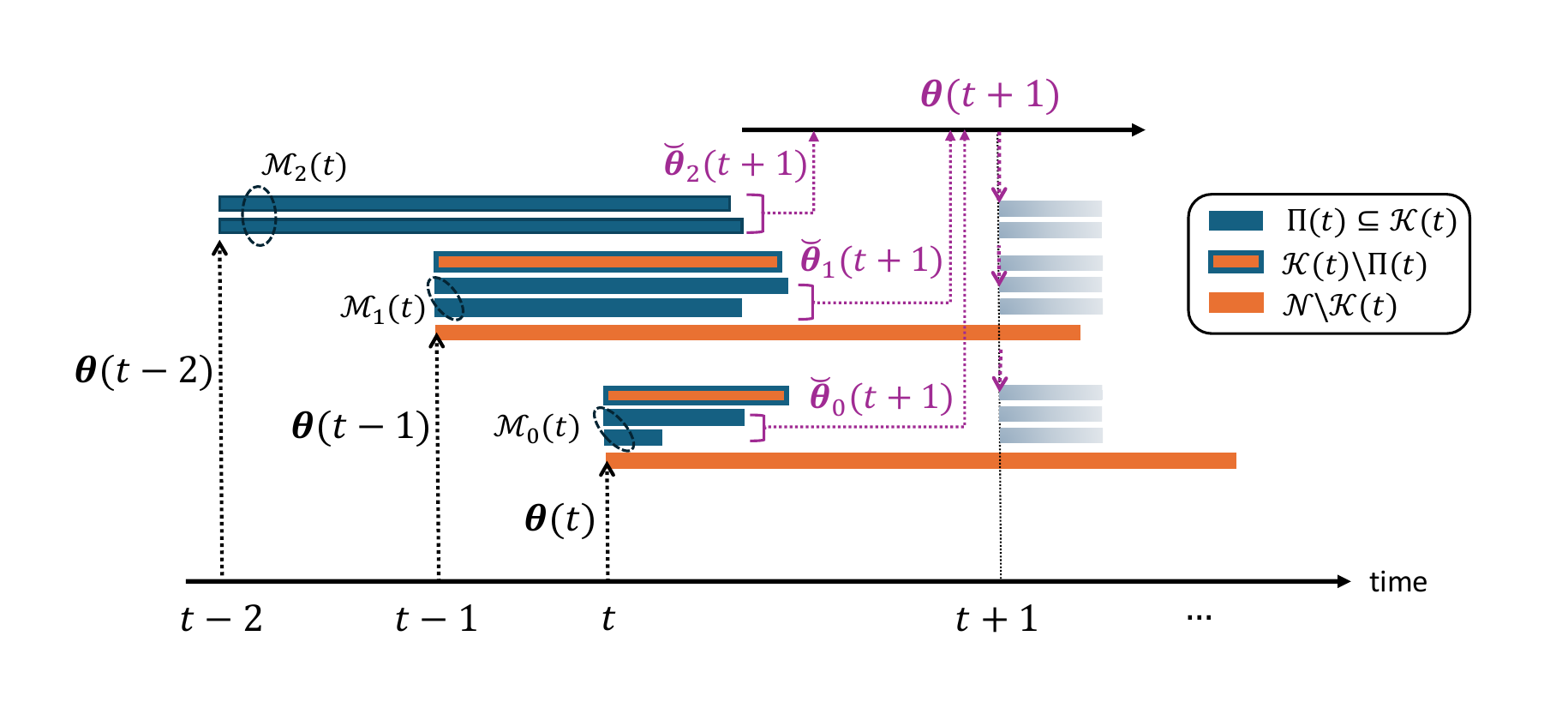}
	\caption{The timelines of $10$ devices from receiving the global model to finishing the local training.
		For iteration $t+1$, the server schedules the subset $\Pi(t)$ from the ready-to-update set $\mathcal{K}(t)$ for the gradient updates, where $\Pi(t)=\cup_{i=0}^2\mathcal{M}_i(t)$, and the devices in $\mathcal{M}_i(t)$ conduct the local training based on $\boldsymbol{\theta}(t-i)$. The average updated model from $\mathcal{M}_i(t)$ is $\breve{\boldsymbol{\theta}}_i(t+1)$, which altogether contributes to the renewed global model $\boldsymbol{\theta}(t+1)$. Upon receiving $\boldsymbol{\theta}(t+1)$, the devices in $\mathcal{K}(t)$ start a new iteration of training.}
	\label{fig:thetaCmp}
\end{figure*}

We aim to find an analytical bound on $\mathbb{E}\left[\|\boldsymbol{\theta}\left(t+1\right)-\boldsymbol{\theta}^*\|^2\right]$.
Recall that $\boldsymbol{\theta}(t+1)$ is the global model at iteration $t+1$, which is computed based on the updates from the devices in $\Pi(t)=\cup_{m=0,...,a_{\lim}}\mathcal{M}_m(t)$.
A scheduled device \mbox{$k\in\mathcal{M}_m(t)\subseteq\Pi(t)$} conducts model training based on the received model $\boldsymbol{\theta}_k(t,0)=\boldsymbol{\theta}(t-m)$.
(The age of device $k$ is $m$.)
The average updated model from the set $\mathcal{M}_m(t)$ can then be computed as
\begin{equation}
	\breve{\boldsymbol{\theta}}_m(t+1)=\sum_{k\in\mathcal{M}_m(t)}\frac{w_k(t)}{c_m(t)}\boldsymbol{\theta}_k(t,1),\label{eq:aggrSubset}
\end{equation}
where $\boldsymbol{\theta}_k(t,1)=\boldsymbol{\theta}_k(t,0)-\alpha\mathcal{Q}_\nu\left[\nabla \tilde{F}_k(\boldsymbol{\theta}_k(t,0);\mathcal{B}(t))\right]$ is the updated model after local training, and $\nabla\tilde{F}_k(\cdot)$ is the randomly-sparsified gradient.
Define the aggregated gradient updates of $\mathcal{M}_m(t)$ as
\begin{align}
	&\boldsymbol{g}_{m}(t)=\sum_{k\in\mathcal{M}_m(t)}\frac{w_k(t)}{c_m(t)}\mathcal{Q}_\nu\left(\nabla \tilde{F}_k\left(\boldsymbol{\theta}(t-m);\mathcal{B}_k(t)\right)\right).
	\label{def:g1}
\end{align}
Then, \eqref{eq:aggrSubset} can be rewritten as
\begin{align*}
	\breve{\boldsymbol{\theta}}_m\left(t+1\right)=\boldsymbol{\theta}(t-m)-\alpha\boldsymbol{g}_{m}(t).
\end{align*}
The aggregated model based on the entire scheduled set, $\Pi(t)$, is then
\begin{equation}
	\boldsymbol{\theta}\left(t+1\right)=\sum_{m=0}^{a_{\lim}}c_m\left(t\right)\breve{\boldsymbol{\theta}}_m\left(t+1\right).
	\label{eq:redefineAggrModel}
\end{equation}
The relations between $\boldsymbol{\theta}(t+1)$, $\breve{\boldsymbol{\theta}}_m(t+1)$, and $\boldsymbol{\theta}(t-m)$ for a system with $a_{\lim}=2$ are illustrated in Figure \ref{fig:thetaCmp}.

\vspace{.5cm} 
We expand $\mathbb{E}\left[\|\boldsymbol{\theta}\left(t+1\right)-\boldsymbol{\theta}^*\|^2\right]$ using \eqref{eq:redefineAggrModel},  Jensen's inequality,  and the fact that $\sum_{m=0}^{a_{\lim}}c_m=1$:
\begin{align}
	&\mathbb{E}\left[\|\boldsymbol{\theta}\left(t+1\right)-\boldsymbol{\theta}^*\|^2\right]=\mathbb{E}\left[\Big|\Big|\sum_{m=0}^{a_{\lim}}c_m\left(t\right)\breve{\boldsymbol{\theta}}_m\left(t+1\right)-\boldsymbol{\theta}^*\Big|\Big|^2\right]\nonumber\\
	&\leq\sum_{m=0}^{a_{\lim}}c_m(t)\mathbb{E}\left[\|\breve{\boldsymbol{\theta}}_m\left(t+1\right)-\boldsymbol{\theta}^*\|^2\right]\nonumber\\
	&\leq\max_{m=0,...,a_{\lim}}\mathbb{E}\left[\|\breve{\boldsymbol{\theta}}_m\left(t+1\right)-\boldsymbol{\theta}^*\|^2\right].
	\label{ieq:optimalityGap1}
\end{align}
To facilitate analysis, we introduce the following lemma that quantifies how the error evolves in every age group.
\begin{lemma}
	\label{lm:subset}
	Under the conditions stated in \textbf{Theorem 1}, for $m\in\{0,...,a_{\lim}\}$, the following result holds
	\begin{align*}
		&\mathbb{E}\left[\|\breve{\boldsymbol{\theta}}_m\left(t+1\right)-\boldsymbol{\theta}^*\|^2\right]\\
		&\leq\left[1-\frac{\mu\alpha r_{\min}}{d}+\alpha^2\left(2L^2+C_3\right)\right]\mathbb{E}\left[\|\boldsymbol{\theta}(t-m)-\boldsymbol{\theta}^*\|^2\right]\nonumber\\
		&\quad+\zeta L\alpha+\alpha^2\left[\zeta\left(2L^2+C_3\right)+4C_1\left(1+\frac{d}{4\nu^2}\right)\right].
	\end{align*}
\end{lemma}
\noindent\emph{Proof.} See Section \ref{lm:subsetPf}.\hfill$\qed$\vspace{.5cm}

Applying Lemma \ref{lm:subset} to \eqref{ieq:optimalityGap1} gives
\begin{align}
	&\mathbb{E}\left[\|\boldsymbol{\theta}\left(t+1\right)-\boldsymbol{\theta}^*\|^2\right]\leq \left[1-\frac{\mu\alpha r_{\min}}{d}+\alpha^2\left(2L^2+C_3\right)\right]\nonumber\\
	&\quad\quad\quad\quad\quad\quad\quad\quad\quad\quad\cdot\max_{0\leq m\leq a_{\lim},t-m\geq 1}\mathbb{E}\left[\|\boldsymbol{\theta}(t-m)-\boldsymbol{\theta}^*\|^2\right]\nonumber\\
	&+\zeta L\alpha+\alpha^2\left[\zeta\left(2L^2+C_3\right)+4C_1\left(1+\frac{d}{4\nu^2}\right)\right].\nonumber
	\\ 
	& =    \bar{C}\max_{0\leq m\leq a_{\lim},t-m\geq 1}\mathbb{E}\left[\|\boldsymbol{\theta}(t-m)-\boldsymbol{\theta}^*\|^2\right]+\alpha\epsilon, \label{ieq:base_induc}
\end{align}
where we defined $\bar{C}=1-\mu\alpha r_{\min}/d+\alpha^2\left(2L^2+C_3\right)$ (note that $\bar{C}<1$) and 
$\epsilon$ is defined in \eqref{eq:epsilon}.

We are now going to prove by induction that  the following holds for all $t$:
\begin{align}
	\mathbb{E}\left[\|\boldsymbol{\theta}\left(t+1\right)-\boldsymbol{\theta}^*\|^2\right]&\leq\bar{C}^{\lfloor\frac{t-1}{a_{\lim}+1}\rfloor+1}\mathbb{E}\left[\|\boldsymbol{\theta}(1)-\boldsymbol{\theta}^*\|^2\right]\nonumber\\
	&\quad+\alpha\epsilon\sum_{i=0}^{t-1}\bar{C}^i.\label{ieq:claimIeq}
\end{align}
Clearly,  \eqref{ieq:claimIeq} holds for  $t=1$, because \eqref{ieq:base_induc} gives (in this case $m=0$)
\begin{align*}
	\mathbb{E}\left[\|\boldsymbol{\theta}\left(2\right)-\boldsymbol{\theta}^*\|^2\right]&\leq\bar{C}\mathbb{E}\left[\|\boldsymbol{\theta}(1)-\boldsymbol{\theta}^*\|^2\right]+\alpha\epsilon.
\end{align*}
Now assume, as induction hypothesis, that (12) holds for \mbox{$t\le t_0$}. 
Then for $t=t_0+1$ we have, based on \eqref{ieq:base_induc},
\begin{align}
	\label{ieq:tPlus1Case}
	&\mathbb{E}\left[\|\boldsymbol{\theta}\left(t_0+2\right)-\boldsymbol{\theta}^*\|^2\right]\nonumber\\
	&\leq\bar{C}\max_{0\leq m\leq\min\left(a_{\lim},t_0\right)}\mathbb{E}\left[\|\boldsymbol{\theta}(t_0+1-m)-\boldsymbol{\theta}^*\|^2\right]+\alpha\epsilon.
\end{align}
Since the first and the second terms in the bound of \eqref{ieq:claimIeq}, for the case of $t\le t_0$, are decreasing and increasing with $t_0$ respectively, \eqref{ieq:tPlus1Case} can be upper-bounded further as
\begin{align}
	&\mathbb{E}\left[\|\boldsymbol{\theta}\left(t_0+2\right)-\boldsymbol{\theta}^*\|^2\right]\nonumber\\
	&\leq\bar{C}\left\{\bar{C}^{\lfloor\frac{t_0-1-\min\left(a_{\lim},t_0\right)}{a_{\lim}+1}\rfloor+1}\mathbb{E}\left[\|\boldsymbol{\theta}(1)-\boldsymbol{\theta}^*\|^2\right]+\alpha\epsilon\sum_{i=0}^{t_0-1}\bar{C}^i\right\}+\alpha\epsilon\nonumber\\
	&=\bar{C}^{\lfloor\frac{t_0+a_{\lim}-\min\left(a_{\lim},t_0\right)}{a_{\lim}+1}\rfloor+1}\mathbb{E}\left[\|\boldsymbol{\theta}(1)-\boldsymbol{\theta}^*\|^2\right]+\alpha\epsilon\sum_{i=0}^{t_0}\bar{C}^i\nonumber\\
	&=\bar{C}^{\lfloor\frac{\max\left(t_0,a_{\lim}\right)}{a_{\lim}+1}\rfloor+1}\mathbb{E}\left[\|\boldsymbol{\theta}(1)-\boldsymbol{\theta}^*\|^2\right]+\alpha\epsilon\sum_{i=0}^{t_0}\bar{C}^i.\label{ieq:case_tplus2}
\end{align}
Since
\begin{equation*}
	\lfloor\frac{\max\left(t_0,a_{\lim}\right)}{a_{\lim}+1}\rfloor=\begin{cases}
		\lfloor\frac{t_0}{a_{\lim}+1}\rfloor,&t>a_{\lim}\\
		\lfloor\frac{a_{\lim}}{a_{\lim}+1}\rfloor=\lfloor\frac{t_0}{a_{\lim}+1}\rfloor ,&{t_0\leq a_{\lim}} 
	\end{cases},
\end{equation*}
\eqref{ieq:case_tplus2} can be rewritten as
\begin{align*}
	\mathbb{E}\left[\|\boldsymbol{\theta}\left(t_0+2\right)-\boldsymbol{\theta}^*\|^2\right]&\leq\bar{C}^{\lfloor\frac{t_0}{a_{\lim}+1}\rfloor+1}\mathbb{E}\left[\|\boldsymbol{\theta}(1)-\boldsymbol{\theta}^*\|^2\right]\nonumber\\
	&\quad+\alpha\epsilon\sum_{i=0}^{t_0}\bar{C}^i,
\end{align*}
which completes induction step.  With this, \eqref{ieq:claimIeq} follows.
Finally, since $\sum_{i=0}^{t-1}\bar{C}^i\leq1/\left(1-\bar{C}\right)$, \eqref{ieq:claimIeq} can be further upper-bounded by \eqref{ieq:thm_rslt}, which completes the proof of the theorem.

\subsection{Proof of Lemma \ref{lm:subset}}
\label{lm:subsetPf}
Note that the design of the resource allocation guarantees that all  $r_k(t)$ are equal $\forall k\in\Pi(t)$. Hence, we let $r_t=r_k(t)$ in the following derivation.
We define the mean of $\boldsymbol{g}_{m}(t)$ in \eqref{def:g1} conditioned on a realization of $r_t$, $\mathcal{M}_m(t)$, and $\boldsymbol{\theta}(t-m)$ as 
\begin{align}
	&\bar{\boldsymbol{g}}_{m}(t)\nonumber\\
	&=\mathbb{E}_{\mathcal{B}_k(t)}\mathbb{E}_{\Psi(t)}\left[\boldsymbol{g}_{m}(t)\right]=\frac{r_t}{d}\sum_{k\in\mathcal{M}_m(t)}\frac{w_k(t)}{c_m(t)}\nabla F_k\left(\boldsymbol{\theta}(t-m)\right),
	\label{def:g_bar1}
\end{align}
where $\Psi(t)=\{\mathcal{Q}_\nu,\mathcal{V}_k(t),\forall k\in\mathcal{M}_m(t)\}$ contains all sources of randomness in the compression operation. 

We rewrite $\|\breve{\boldsymbol{\theta}}_m\left(t+1\right)-\boldsymbol{\theta}^*\|^2$ in terms of the received model $\boldsymbol{\theta}(t-m)$, using \eqref{def:g1} and \eqref{def:g_bar1}, and  expand the $\|\cdot\|^2$:
\begin{align}
	&\|\breve{\boldsymbol{\theta}}_m\left(t+1\right)-\boldsymbol{\theta}^*\|^2\nonumber\\
	&=\|\boldsymbol{\theta}(t-m)-\alpha\boldsymbol{g}_{m}(t)-\boldsymbol{\theta}^*\|^2\nonumber\\
	&=\|\boldsymbol{\theta}(t-m)-\alpha\boldsymbol{g}_{m}(t)-\boldsymbol{\theta}^*-\alpha\bar{\boldsymbol{g}}_{m}(t)+\alpha\bar{\boldsymbol{g}}_{m}(t)\|^2\nonumber\\
	&=\|\boldsymbol{\theta}(t-m)-\boldsymbol{\theta}^*-\alpha\bar{\boldsymbol{g}}_{m}(t)\|^2+\alpha^2 \|\boldsymbol{g}_{m}(t)-\bar{\boldsymbol{g}}_{m}(t)\|^2\nonumber\\
	&\quad-2\alpha\left[\left(\boldsymbol{\theta}(t-m)-\boldsymbol{\theta}^*-\alpha\bar{\boldsymbol{g}}_{m}(t)\right)^T\left(\boldsymbol{g}_{m}(t)-\bar{\boldsymbol{g}}_{m}(t)\right)\right].
	\label{eq:theta_diff}
\end{align}
Then, the first term in \eqref{eq:theta_diff} can be bounded by the following Lemma:
\begin{lemma}
	Under the conditions stated in \textbf{Theorem 1},
	\begin{align*}
		&\|\boldsymbol{\theta}(t-m)-\boldsymbol{\theta}^*-\alpha\bar{\boldsymbol{g}}_{m}(t)\|^2\nonumber\\
		&\leq\left(1-\frac{\mu\alpha r_{\min}}{d}+2L^2\alpha^2\right)\|\boldsymbol{\theta}(t-m)-\boldsymbol{\theta}^*\|^2+\zeta L\alpha\left(2L\alpha+1\right).
	\end{align*}
	\label{lm:rearrange}
\end{lemma}
\noindent\emph{Proof.} See Section \ref{sec:pf_lm_rearrange}.\hfill$\qed$\vspace{.5cm}

Then, the total expectation of \eqref{eq:theta_diff} can be computed as
\begin{align}
	&\mathbb{E}\left[\|\breve{\boldsymbol{\theta}}_m\left(t+1\right)-\boldsymbol{\theta}^*\|^2\right]\nonumber\\
	&\leq\left(1-\frac{\mu\alpha r_{\min}}{d}+2L^2\alpha^2\right)\mathbb{E}\left[\|\boldsymbol{\theta}(t-m)-\boldsymbol{\theta}^*\|^2\right]\nonumber\\
	&\quad+\alpha^2 \mathbb{E}\left[\|\boldsymbol{g}_{m}(t)-\bar{\boldsymbol{g}}_{m}(t)\|^2\right]+\zeta L\alpha\left(2L\alpha+1\right)\nonumber\\
	&\quad-2\alpha \mathbb{E}\left[\left(\boldsymbol{\theta}(t-m)-\boldsymbol{\theta}^*-\alpha \bar{\boldsymbol{g}}_{m}(t)\right)^T\left(\boldsymbol{g}_{m}(t)-\bar{\boldsymbol{g}}_{m}(t)\right)\right].
	\label{eq:theta_diff_2}
\end{align}
The second term in \eqref{eq:theta_diff_2},  $\mathbb{E}\left[\|\boldsymbol{g}_{m}(t)-\bar{\boldsymbol{g}}_{m}(t)\|^2\right]$,  can be bounded by the following Lemma:
\begin{lemma}
	\label{lm:var_g}
	Under the condition stated in \textbf{Theorem 1}, the variance of $\boldsymbol{g}_{m}(t)$ fulfills
	\begin{align*}
		\mathbb{E}\left[\|\boldsymbol{g}_{m}(t)-\bar{\boldsymbol{g}}_{m}(t)\|^2\right]&\leq C_3\mathbb{E}\left[\|\boldsymbol{\theta}(t-m)-\boldsymbol{\theta}^*\|^2\right]+C_3\zeta\nonumber\\
		&\quad+4C_1\left(1+\frac{d}{4\nu^2}\right)
	\end{align*}
\end{lemma}
\noindent\emph{Proof.} See Section \ref{sec:pf_lemma2}.\hfill$\qed$\vspace{.5cm}

With $\bar{\boldsymbol{g}}_{m}(t)$ defined as in \eqref{def:g_bar1}, 
it holds that 
\begin{equation}
	\mathbb{E}_{\mathcal{B}_k(t)}\mathbb{E}_{\Psi(t)}\left[\left(\boldsymbol{\theta}(t-m)-\boldsymbol{\theta}^*-\alpha \bar{\boldsymbol{g}}_{m}(t)\right)^T\left(\boldsymbol{g}_{m}(t)-\bar{\boldsymbol{g}}_{m}(t)\right)\right]=0;\label{eq:last_term_0}
\end{equation}
therefore, by taking total expectation, the fourth term of  \eqref{eq:theta_diff_2} is zero.
Using  \textbf{Lemma \ref{lm:var_g}} and \eqref{eq:last_term_0} in \eqref{eq:theta_diff_2} completes the proof.

\subsection{Proof of Lemma \ref{lm:rearrange}}
\label{sec:pf_lm_rearrange}
We insert $\bar{\boldsymbol{g}}_{m}(t)$ in \eqref{def:g_bar1} into 
$\|\boldsymbol{\theta}(t-m)-\boldsymbol{\theta}^*-\alpha \bar{\boldsymbol{g}}_{m}(t)\|^2$ 
and expand the squared norm:
\begin{align}
	&\|\boldsymbol{\theta}(t-m)-\boldsymbol{\theta}^*-\alpha \bar{\boldsymbol{g}}_{m}(t)\|^2\nonumber\\
	&=\|\boldsymbol{\theta}(t-m)-\boldsymbol{\theta}^*\|^2+\alpha^2 \left\|\frac{r_t}{d}\sum_{k\in\mathcal{M}_m(t)}\frac{w_k(t)}{c_m(t)}\nabla F_k\left(\boldsymbol{\theta}(t-m)\right)\right\|^2\nonumber\\
	&\quad-\frac{2\alpha r_t}{d}\sum_{k\in\mathcal{M}_m(t)}\frac{w_k(t)}{c_m(t)}\left(\boldsymbol{\theta}(t-m)-\boldsymbol{\theta}^*\right)^T\nabla F_k\left(\boldsymbol{\theta}(t-m)\right)\nonumber\\
	&\leq\|\boldsymbol{\theta}(t-m)-\boldsymbol{\theta}^*\|^2+\frac{r_t^2\alpha^2 }{d^2}\sum_{k\in\mathcal{M}_m(t)}\frac{w_k(t)}{c_m(t)}\|\nabla F_k\left(\boldsymbol{\theta}(t-m)\right)\|^2\nonumber\\
	&\quad-\frac{2\alpha r_t}{d}\sum_{k\in\mathcal{M}_m(t)}\frac{w_k(t)}{c_m(t)}\left(\boldsymbol{\theta}(t-m)-\boldsymbol{\theta}^*\right)^T\nabla F_k\left(\boldsymbol{\theta}(t-m)\right),\label{eq:A_3_async_1}
\end{align}
where \eqref{eq:A_3_async_1} follows from Jensen's inequality (the normalized weights sum up to $1$).
The third term in \eqref{eq:A_3_async_1} is upper bounded by the strong convexity property of $F_k(\cdot)$, 
\begin{align}
	&-\left(\boldsymbol{\theta}(t-m)-\boldsymbol{\theta}^*\right)^T\nabla F_k\left(\boldsymbol{\theta}(t-m)\right)\nonumber\\
	&\leq-\left(F_k\left(\boldsymbol{\theta}(t-m)\right)-F_k\left(\boldsymbol{\theta}^*\right)\right)-\frac{\mu}{2}\|\boldsymbol{\theta}(t-m)-\boldsymbol{\theta}^*\|^2.
	\label{eq:A_3_2_async}
\end{align}
By using $L$-smoothness on the second term of  \eqref{eq:A_3_async_1}, and
using (\ref{eq:A_3_2_async}) on the third term of \eqref{eq:A_3_async_1},
Eq. \eqref{eq:A_3_async_1} becomes
\begin{align}
	&\|\boldsymbol{\theta}(t-m)-\boldsymbol{\theta}^*-\alpha \bar{\boldsymbol{g}}_{m}(t)\|^2\nonumber\\
	&\leq\|\boldsymbol{\theta}(t-m)-\boldsymbol{\theta}^*\|^2+\frac{L^2r_t^2\alpha^2 }{d^2}\sum_{k\in\mathcal{M}_m(t)}\frac{w_k(t)}{c_m(t)}\|\boldsymbol{\theta}(t-m)-\boldsymbol{\theta}_k^*\|^2\nonumber\\
	&\quad-\frac{2\alpha r_t}{d}\sum_{k\in\mathcal{M}_m(t)}\frac{w_k(t)}{c_m(t)}\Big[F_k\left(\boldsymbol{\theta}(t-m)\right)-F_k\left(\boldsymbol{\theta}^*\right)\nonumber\\
	&\quad\quad\quad\quad\quad\quad\quad\quad\quad\quad\quad+\frac{\mu}{2}\|\boldsymbol{\theta}(t-m)-\boldsymbol{\theta}^*\|^2\Big].\label{ieq:cmb_terms}
\end{align}
By collecting  the terms in (\ref{ieq:cmb_terms}) containing $\|\boldsymbol{\theta}(t-m)-\boldsymbol{\theta}^*\|^2$,
writing $\boldsymbol{\theta}(t-m)-\boldsymbol{\theta}^*_k = \boldsymbol{\theta}(t-m)-\boldsymbol{\theta}^* + \boldsymbol{\theta}^* - \boldsymbol{\theta}^*_k$,  
and applying the inequality
\begin{equation}
	\|\boldsymbol{a}+\boldsymbol{b}\|^2\leq 2\|\boldsymbol{a}\|^2+2\|\boldsymbol{b}\|^2,\label{ieq:norm2_ieq}
\end{equation}
\eqref{ieq:cmb_terms} can be rewritten as
\begin{align}
	&\|\boldsymbol{\theta}(t-m)-\boldsymbol{\theta}^*-\alpha \bar{\boldsymbol{g}}_{m}(t)\|^2\nonumber\\
	&\leq\left(1-\frac{\mu\alpha r_t}{d}\right)\|\boldsymbol{\theta}(t-m)-\boldsymbol{\theta}^*\|^2\nonumber\\
	&\quad+\frac{2L^2r_t^2\alpha^2 }{d^2}\left[\|\boldsymbol{\theta}(t-m)-\boldsymbol{\theta}^*\|^2+\sum_{k\in\mathcal{M}_m(t)}\frac{w_k(t)}{c_m(t)}\|\boldsymbol{\theta}^*-\boldsymbol{\theta}_k^*\|^2\right]\nonumber\\
	&\quad+\frac{2\alpha r_t}{d}\sum_{k\in\mathcal{M}_m(t)}\frac{w_k(t)}{c_m(t)}\Big[F_k\left(\boldsymbol{\theta}^*\right)-F_k\left(\boldsymbol{\theta}(t-m)\right)\Big]\nonumber\\
	&\leq\left(1-\frac{\mu\alpha r_t}{d}+\frac{2L^2r_t^2\alpha^2 }{d^2}\right)\|\boldsymbol{\theta}(t-m)-\boldsymbol{\theta}^*\|^2\nonumber\\
	&\quad+\frac{2L^2r_t^2\alpha^2 }{d^2}\sum_{k\in\mathcal{M}_m(t)}\frac{w_k(t)}{c_m(t)}\|\boldsymbol{\theta}^*-\boldsymbol{\theta}_k^*\|^2\nonumber\\
	&\quad+\frac{2\alpha r_t}{d}\sum_{k\in\mathcal{M}_m(t)}\frac{w_k(t)}{c_m(t)}\Big[F_k\left(\boldsymbol{\theta}^*\right)-F_k^*\Big],\label{ieq:new_method}
\end{align}
where $F_k\left(\boldsymbol{\theta}(t-m)\right)\geq F_k^*$ is used in the last inequality. By using $L$-smoothness, the sum in the last term of (\ref{ieq:new_method}) can be bounded as
\begin{equation}
	\sum_{k\in\mathcal{M}_m(t)}\frac{w_k(t)}{c_m(t)}\left[F_k\left(\boldsymbol{\theta}^*\right)-F_k^*\right]\leq\frac{L}{2}\sum_{k\in\mathcal{M}_m(t)}\frac{w_k(t)}{c_m(t)}\|\boldsymbol{\theta}^*-\boldsymbol{\theta}_k^*\|^2.\label{ieq:applymuL}
\end{equation}
Then, by using \eqref{ieq:applymuL} and \eqref{eq:noniid_b1} in \eqref{ieq:new_method}, we have
\begin{align}
	&\|\boldsymbol{\theta}(t-m)-\boldsymbol{\theta}^*-\alpha \bar{\boldsymbol{g}}_{m}(t)\|^2\nonumber\\
	&\leq\left(1-\frac{\mu\alpha r_t}{d}+\frac{2L^2r_t^2\alpha^2 }{d^2}\right)\|\boldsymbol{\theta}(t-m)-\boldsymbol{\theta}^*\|^2\nonumber\\
	&\quad+\frac{r_t\zeta L\alpha}{d}\left(1+\frac{2Lr_t\alpha}{d}\right).\label{ieq:lmfirstfnl}
\end{align}
The proof is complete by further upper-bounding \eqref{ieq:lmfirstfnl} with
$r_t/d\leq 1$ and using \eqref{ieq:rmin}.

\subsection{Proof of Lemma \ref{lm:var_g}}
\label{sec:pf_lemma2}
The variance of $\boldsymbol{g}_m(t)$ with respect to the randomness of the mini-batch $\mathcal{B}_k(t), \forall k\in\mathcal{M}_m(t)$, and compression operation $\Psi(t)$, conditioned on a realization of $r_t$, $\mathcal{M}_m(t)$, and $\boldsymbol{\theta}(t-m)$, can be expressed as 
\begin{align}
	&\mathbb{E}_{\mathcal{B}_k(t)}\mathbb{E}_{\Psi(t)}\left[\|\boldsymbol{g}_{m}(t)-\bar{\boldsymbol{g}}_{m}(t)\|^2\right]\nonumber\\
	&=\mathbb{E}_{\mathcal{B}_k(t)}\mathbb{E}_{\Psi(t)}\Bigg[\Big\|\sum_{k\in\mathcal{M}_m(t)}\frac{w_k(t)}{c_m(t)}\Big[\mathcal{Q}_\nu\left(\nabla \tilde{F}_k\left(\boldsymbol{\theta}(t-m);\mathcal{B}_k(t)\right)\right)\nonumber\\
	&\quad\quad\quad\quad\quad\quad\quad\quad\quad\quad\quad\quad\quad\quad-\frac{r_t}{d}\nabla F_k\left(\boldsymbol{\theta}(t-m)\right)\Big]\Big\|^2\Bigg].\label{eq:g_var}
\end{align}
Based on convexity of $\|\cdot\|^2$ and the inequality in \eqref{ieq:norm2_ieq}, \eqref{eq:g_var} can be bounded by
\begin{align}
	&\mathbb{E}_{\mathcal{B}_k(t)}\mathbb{E}_{\Psi(t)}\left[\|\boldsymbol{g}_{m}(t)-\bar{\boldsymbol{g}}_{m}(t)\|^2\right]\nonumber\\
	&\leq\sum_{k\in\mathcal{M}_m(t)}\frac{w_k(t)}{c_m(t)}\mathbb{E}_{\mathcal{B}_k(t)}\mathbb{E}_{\Psi(t)}\Big[\|\mathcal{Q}_\nu\left(\nabla \tilde{F}_k\left(\boldsymbol{\theta}(t-m);\mathcal{B}_k(t)\right)\right)\nonumber\\
	&\quad\quad\quad\quad\quad\quad\quad\quad\quad\quad\quad\quad\quad-\frac{r_t}{d}\nabla F_k\left(\boldsymbol{\theta}(t-m)\right)\|^2\Big]\nonumber\\
	&\leq 2\sum_{k\in\mathcal{M}_m(t)}\frac{w_k(t)}{c_m(t)}\mathbb{E}_{\mathcal{B}_k(t)}\mathbb{E}_{\mathcal{Q}_\nu}\mathbb{E}_{\mathcal{V}_k(t)}\big[\nonumber\\
	&\quad\quad\quad\quad\quad\|\mathcal{Q}_\nu\left(\nabla \tilde{F}_k\left(\boldsymbol{\theta}(t-m);\mathcal{B}_k(t)\right)\right)-\nabla F_k\left(\boldsymbol{\theta}(t-m)\right)\|^2\big]\nonumber\\
	&\quad+2\sum_{k\in\mathcal{M}_m(t)}\frac{w_k(t)}{c_m(t)}\|\nabla F_k\left(\boldsymbol{\theta}(t-m)\right)-\frac{r_t}{d}\nabla F_k\left(\boldsymbol{\theta}(t-m)\right)\|^2,\label{ieq:compExpand}
\end{align}
where the compression operation $\Psi(t)$ consists of  sparsification $\mathcal{V}_k(t)$ followed by quantization $\mathcal{Q}_{\nu}$ (both are mappings with randomness).
Note that $\mathbb{E}_{\mathcal{V}_k(t)}\small[\|\mathcal{Q}_\nu\left(\nabla \tilde{F}_k\left(\boldsymbol{\theta}(t-m);\mathcal{B}_k(t)\right)\right)-\nabla F_k\left(\boldsymbol{\theta}(t-m)\right)\|^2\small]$ in \eqref{ieq:compExpand} is an expected value w.r.t. the randomness in  sparsification operation and can be computed as
\begin{align}
	&\mathbb{E}_{\mathcal{V}_k(t)}\small[\|\mathcal{Q}_\nu\left(\nabla \tilde{F}_k\left(\boldsymbol{\theta}(t-m);\mathcal{B}_k(t)\right)\right)-\nabla F_k\left(\boldsymbol{\theta}(t-m)\right)\|^2\small]\nonumber\\
	&=\frac{r_t}{d}\|\mathcal{Q}_\nu\left(\nabla F_k\left(\boldsymbol{\theta}(t-m);\mathcal{B}_k(t)\right)\right)-\nabla F_k\left(\boldsymbol{\theta}(t-m)\right)\|^2\nonumber\\
	&\quad+\frac{d-r_t}{d}\|\nabla F_k\left(\boldsymbol{\theta}(t-m)\right)\|^2,\label{eq:after_rdmSp}
\end{align}
because each element in $\nabla \tilde{F}_k\left(\boldsymbol{\theta}(t-m);\mathcal{B}_k(t)\right)$ has probability $r_t/d$ to be preserved and probability $1-r_t/d$ to be nullified.
By inserting \eqref{eq:after_rdmSp} into \eqref{ieq:compExpand},
\begin{align}
	&\mathbb{E}_{\mathcal{B}_k(t)}\mathbb{E}_{\Psi(t)}\left[\|\boldsymbol{g}_{m}(t)-\bar{\boldsymbol{g}}_{m}(t)\|^2\right]\nonumber\\
	&\leq2\sum_{k\in\mathcal{M}_m(t)}\frac{w_k(t)}{c_m(t)}\mathbb{E}_{\mathcal{B}_k(t)}\mathbb{E}_{\mathcal{Q}_\nu}\Big[\frac{d-r_t}{d}\|\nabla F_k\left(\boldsymbol{\theta}(t-m)\right)\|^2\nonumber\\
	&\quad\quad\quad+\frac{r_t}{d}\|\mathcal{Q}_\nu\left(\nabla F_k\left(\boldsymbol{\theta}(t-m);\mathcal{B}_k(t)\right)\right)-\nabla F_k\left(\boldsymbol{\theta}(t-m)\right)\|^2\Big]\nonumber\\
	&\quad+2\sum_{k\in\mathcal{M}_m(t)}\frac{w_k(t)}{c_m(t)}\left(\frac{d-r_t}{d}\right)^2\|\nabla F_k\left(\boldsymbol{\theta}(t-m)\right)\|^2.\label{ieq:after_spar}
\end{align}
By further using the fact that $\left(\frac{d-r_t}{d}\right)^2\leq\frac{d-r_t}{d}\leq 1$ and applying \eqref{ieq:norm2_ieq} again, \eqref{ieq:after_spar} leads to
\begin{align*}
	&\mathbb{E}_{\mathcal{B}_k(t)}\mathbb{E}_{\Psi(t)}\left[\|\boldsymbol{g}_{m}(t)-\bar{\boldsymbol{g}}_{m}(t)\|^2\right]\nonumber\\
	&\leq\frac{4r_t}{d}\sum_{k\in\mathcal{M}_m(t)}\frac{w_k(t)}{c_m(t)}\mathbb{E}_{\mathcal{B}_k(t)}\mathbb{E}_{\mathcal{Q}_\nu}\Big[\nonumber\\
	&\quad\|\nabla F_k\left(\boldsymbol{\theta}(t-m);\mathcal{B}_k(t)\right)-\nabla F_k\left(\boldsymbol{\theta}(t-m)\right)\|^2\nonumber\\
	&\quad+\|\mathcal{Q}_\nu\left(\nabla F_k\left(\boldsymbol{\theta}(t-m);\mathcal{B}_k(t)\right)\right)-\nabla F_k\left(\boldsymbol{\theta}(t-m);\mathcal{B}_k(t)\right)\|^2\Big]\nonumber\\
	&\quad+4\sum_{k\in\mathcal{M}_m(t)}\frac{w_k(t)}{c_m(t)}\|\nabla F_k\left(\boldsymbol{\theta}(t-m)\right)\|^2.
\end{align*}
We then evaluate the expectation over the random quantizer, i.e., $\mathbb{E}_{\mathcal{Q}_{\nu}}[\cdot]$, to obtain
\begin{align}
	&\mathbb{E}_{\mathcal{B}_k(t)}\mathbb{E}_{\Psi(t)}\left[\|\boldsymbol{g}_{m}(t)-\bar{\boldsymbol{g}}_{m}(t)\|^2\right]\nonumber\\
	&\leq\frac{4r_t}{d}\sum_{k\in\mathcal{M}_m(t)}\frac{w_k(t)}{c_m(t)}\mathbb{E}_{\mathcal{B}_k(t)}\Big[\frac{d\|\nabla F_k\left(\boldsymbol{\theta}(t-m);\mathcal{B}_k(t)\right)\|^2}{4\nu^2}\nonumber\\
	&\quad\quad\quad\quad\quad\quad+\|\nabla F_k\left(\boldsymbol{\theta}(t-m);\mathcal{B}_k(t)\right)-\nabla F_k\left(\boldsymbol{\theta}(t-m)\right)\|^2\Big]\nonumber\\
	&\quad+4\sum_{k\in\mathcal{M}_m(t)}\frac{w_k(t)}{c_m(t)}\|\nabla F_k\left(\boldsymbol{\theta}(t-m)\right)\|^2.\label{ieq:after_expt_qnt}
\end{align}
Finally, we evaluate the expectation over the mini-batch sampling $\mathcal{B}_k(t), \forall k\in\mathcal{M}_m(t)$, by using Assumption \ref{asump:sgdCt} in \eqref{ieq:after_expt_qnt},
\begin{align}
	&\mathbb{E}_{\mathcal{B}_k(t)}\mathbb{E}_{\Psi(t)}\left[\|\boldsymbol{g}_{m}(t)-\bar{\boldsymbol{g}}_{m}(t)\|^2\right]\nonumber\\
	&\leq\frac{4r_t}{d}\left(1+\frac{d}{4\nu^2}\right)\left(C_1+C_2\sum_{k\in\mathcal{M}_m(t)}\frac{w_k(t)}{c_m(t)}\|\nabla F_k(\boldsymbol{\theta}(t-m))\|^2\right)\nonumber\\
	&\quad+4\sum_{k\in\mathcal{M}_m(t)}\frac{w_k(t)}{c_m(t)}\|\nabla F_k\left(\boldsymbol{\theta}(t-m)\right)\|^2\nonumber\\
	&\leq\frac{4r_t}{d}\left(1+\frac{d}{4\nu^2}\right)\left[C_1+L^2C_2\sum_{k\in\mathcal{M}_m(t)}\frac{w_k(t)}{c_m(t)}\|\boldsymbol{\theta}(t-m)-\boldsymbol{\theta}_k^*\|^2\right]\nonumber\\
	&\quad+4L^2\sum_{k\in\mathcal{M}_m(t)}\frac{w_k(t)}{c_m(t)}\|\boldsymbol{\theta}(t-m)-\boldsymbol{\theta}_k^*\|^2,\label{eq:proofLm2}
\end{align}
where in the last inequality we used the $L$-smoothness property: $\|\nabla F_k(\boldsymbol{\theta}(t-m))\|^2\leq L^2\|\boldsymbol{\theta}(t-m)-\boldsymbol{\theta}_k^*\|^2$.  
Note that using, again, \eqref{ieq:norm2_ieq},
\begin{align}
	&\sum_{k\in\mathcal{M}_m(t)}\frac{w_k(t)}{c_m(t)}\|\boldsymbol{\theta}(t-m)-\boldsymbol{\theta}_k^*\|^2\nonumber\\
	&\leq2\sum_{k\in\mathcal{M}_m(t)}\frac{w_k(t)}{c_m(t)}\left(\|\boldsymbol{\theta}(t-m)-\boldsymbol{\theta}^*\|^2+\| \boldsymbol{\theta}^*-\boldsymbol{\theta}_k^*\|^2\right)\nonumber\\
	&\leq2\left(\|\boldsymbol{\theta}(t-m)-\boldsymbol{\theta}^*\|^2+\zeta\right).
	\label{eq:proofLm2_1}
\end{align}
By using the fact that $r_t/d\leq 1$ and inserting \eqref{eq:proofLm2_1} in \eqref{eq:proofLm2}, we have
\begin{align*}
	&\mathbb{E}_{\mathcal{B}_k(t)}\mathbb{E}_{\Psi(t)}\left[\|\boldsymbol{g}_{m}(t)-\bar{\boldsymbol{g}}_{m}(t)\|^2\right]\nonumber\\
	&\leq\frac{4r_t}{d}\left(1+\frac{d}{4\nu^2}\right)\left[C_1+2L^2C_2\left(\|\boldsymbol{\theta}(t-m)-\boldsymbol{\theta}^*\|^2+\zeta\right)\right]\nonumber\\
	&\quad+8L^2\left(\|\boldsymbol{\theta}(t-m)-\boldsymbol{\theta}^*\|^2+\zeta\right)\nonumber\\
	&\leq 8L^2\left[\left(1+\frac{d}{4\nu^2}\right)C_2+1\right]\|\boldsymbol{\theta}(t-m)-\boldsymbol{\theta}^*\|^2+4C_1\left(1+\frac{d}{4\nu^2}\right)\nonumber\\
	&\quad+8L^2\zeta\left[C_2\left(1+\frac{d}{4\nu^2}\right)+1\right]\nonumber\\
	&= C_3\|\boldsymbol{\theta}(t-m)-\boldsymbol{\theta}^*\|^2+C_3\zeta+4C_1\left(1+\frac{d}{4\nu^2}\right).
\end{align*}

\bibliographystyle{IEEEtran}
\bibliography{ref-FL.bib}
	
\end{document}